\documentclass[times,twocolumn,final]{elsarticle}

\usepackage{medima}
\usepackage{framed,multirow}
\usepackage{bm}
\usepackage{amsmath}
\usepackage{ulem} % ----

\usepackage{graphicx} %插入图片的宏包
\usepackage{float} %设置图片浮动位置的宏包
\usepackage{subfigure} %插入多图时用子图显示的宏包
\usepackage{stfloats}
\usepackage[table]{xcolor}
\usepackage{arydshln}

\usepackage{appendix}
\usepackage{subfigure}
\usepackage[]{caption2} %新增调用的宏包
\renewcommand{\thesubfigure}{(\roman{subfigure})}%此外，还可设置图编号显示格式，加括号或者不加括号
\makeatletter \renewcommand{\@thesubfigure}{\thesubfigure \space}%子图编号与名称的间隔设置
\renewcommand{\p@subfigure}{} \makeatother

\usepackage{booktabs}

\usepackage[linesnumbered,ruled,vlined]{algorithm2e}

\usepackage{amssymb}
\usepackage{latexsym}
\usepackage{makecell}

\usepackage{url}
\usepackage{xcolor}

\usepackage{hyperref} 

\definecolor{newcolor}{rgb}{.8,.349,.1}

\journal{XXXX}

\begin{document}

\verso{Fei Kong \textit{et~al.}}   

\begin{frontmatter}

\title{Federated attention consistent learning models for prostate cancer diagnosis and Gleason grading}

% \tnotetext[tnote1]{This is an example of title footnote coding.}

\author[1]{Fei Kong\fnref{fn1}}
\fntext[fn1]{These authors contributed equally to this paper.}

% \author[1]{Given-name2 \snm{Surname2}\fnref{fn1}}
\author[2]{Xiyue Wang \fnref{fn1}} %
\author[3]{Jinxi Xiang \fnref{fn1}} % 
\author[3] {Sen Yang} % 

%% Third author's email
\author[4]{Xinran Wang} % Xinran Wang, Meng Yue, Yueping Liu
\author[4]{Meng Yue} %

\author[3]{Jun Zhang\corref{cor1}}
% \cortext[cor1]{Corresponding author: 
%   Tel.: +0-000-000-0000;  
%   fax: +0-000-000-0000;}
\author[5,6,7]{Junhan Zhao} %
\author[3]{Xiao Han} %
\author[1]{Yuhan Dong} %
\author[8]{Biyue Zhu}  % 
\author[9]{Fang Wang\corref{cor1}} %
\author[4]{Yueping Liu\corref{cor1}} %

% \cortext[cor1]{Corresponding author: junejzhang@tencent.com; 578563666@qq.com; liuyp@hebmu.edu.cn;}

\address[1]{Shenzhen International Graduate School, Tsinghua University, Shenzhen, 518055, China}
\address[2]{College of Biomedical Engineering, Sichuan University, Chengdu, 610065, China}
\address[3]{AI Lab, Tencent, Shenzhen, 518057, China}
\address[4]{Department of Pathology, The Fourth Hospital of Hebei Medical University, Shijiazhuang, 050035, China}
\address[5]{Massachusetts General Hospital, Boston, MA, 02114, United States}
\address[6]{Harvard T.H. Chan School of Public Health, Boston, MA, 02115, United States}
\address[7]{Department of Biomedical Informatics, Harvard Medical School, Boston, MA, 02115, United States}
\address[8]{Department of Pharmacy, Children's Hospital of Chongqing Medical University, Chongqing, 400014, China.}
\address[9]{Department of Pathology, The Affiliated Yantai Yuhuangding Hospital of Qingdao University, Yantai, 264000, China}

% \received{1 May 2013}
% \finalform{10 May 2013}
% \accepted{13 May 2013}
% \availableonline{15 May 2013}
% \communicated{S. Sarkar}

\begin{abstract}

Artificial intelligence (AI) holds significant promise in transforming medical imaging, enhancing diagnostics, and refining treatment strategies. However, the reliance on extensive multicenter datasets for training AI models poses challenges due to privacy concerns. Federated learning provides a solution by facilitating collaborative model training across multiple centers without sharing raw data. This study introduces a federated attention-consistent learning (FACL) framework to address challenges associated with large-scale pathological images and data heterogeneity. FACL enhances model generalization by maximizing attention consistency between local clients and the server model. To ensure privacy and validate robustness, we incorporated differential privacy by introducing noise during parameter transfer. We assessed the effectiveness of FACL in cancer diagnosis and Gleason grading tasks using 19,461 whole-slide images of prostate cancer from multiple centers. In the diagnosis task, FACL achieved an area under the curve (AUC) of 0.9718, outperforming seven centers with an average AUC of 0.9499 when categories are relatively balanced. For the Gleason grading task, FACL attained a Kappa score of 0.8463, surpassing the average Kappa score of 0.7379 from six centers. In conclusion, FACL offers a robust, accurate, and cost-effective AI training model for prostate cancer pathology while maintaining effective data safeguards.

\end{abstract}

\end{frontmatter}
% =================================================================================
% ============================== I. INTRODUCTION ==================================
% =================================================================================
\section{Introduction}

Prostate cancer ranks as the second most prevalent cancer and the fifth leading cause of cancer-related deaths among men \cite{sung2021global}. The standard approach for diagnosing solid tumors involves examining hematoxylin and eosin (H\&E)--stained whole-slide images (WSIs) \cite{lu2015automated, schirris2022deepsmile, chikontwe2022feature}. Accurate diagnosis and Gleason grading of prostate cancer WSIs are crucial for effective clinical management and treatment \cite{kartasalo2021artificial, ma2023identification}. Gleason grading, in particular, is the most reliable method for assessing aggressiveness \cite{delahunt2012gleason, silva2020going, short2019gleason}. However, interobserver and intraobserver variability in Gleason scores can result in under or over-treatment of patients in real-world scenarios \cite{kartasalo2021artificial, GEORGE2022262}. Moreover, manually labeling pathological images for diagnosis is time-consuming and expensive \cite{campanella2019clinical, wang2022sclwc, wang2023retccl, xiang2023automatic}.

Recent advancements in artificial intelligence (AI) have accelerated its application in pathological image diagnosis. Groundbreaking research studies have demonstrated AI's ability to deliver precise, cost-effective, and scalable solutions by automating patient data analysis \cite{kartasalo2021artificial, bulten2022artificial, bulten2020automated, takahashi2023artificial}. Numerous prior studies have unequivocally confirmed the effectiveness of AI models in diagnosing prostate cancer and conducting Gleason grading tasks  \cite{campanella2019clinical, bulten2020automated, perincheri2021independent, nagpal2020development, pantanowitz2020artificial, mun2021yet}. However, it is essential to recognize that these AI models demand a significant volume of training data to achieve optimal performance \cite{muti2021development, echle2020clinical, saldanha2022swarm}. Consequently, a common approach involves gathering data from various centers, including medical institutions and hospitals, and then training AI models through centralized learning processes. Nevertheless, it is critical to acknowledge that transferring extensive amounts of sensitive patient data among clinics, hospitals, and other medical establishments may jeopardize patient privacy and potentially conflict with data protection regulations. Additionally, there are notable technical challenges to surmount, such as ensuring efficient data transfer and addressing storage requirements, which pose significant obstacles in this pursuit \cite{saldanha2022swarm, pati2022federated}.

Federated learning (FL) presents an innovative paradigm wherein models are trained by sharing model parameter updates from decentralized data sources \cite{pati2022federated, liu2021federated, li2020federated, xu2021federated, sheller2020federated}. Local retention of data effectively addresses issues related to excessive data transmission and ownership \cite{rieke2020future}. By directly sharing model parameters rather than private data from multiple centers, FL offers a promising solution for applications involving sensitive data. FL training methods generally fall into two categories: (1) primary server and (2) peer-to-peer \cite{lu2022federated}, with the key distinction being whether a primary server is employed to collect model parameters for each client. FL has demonstrated significant potential for sensitive data analytics across various healthcare domains \cite{xu2021federated}, including electronic health records (EHR) \cite{brisimi2018federated, boughorbel2019federated, huang2019patient, min2019predictive, duan2020learning}, healthcare Internet of Things (IoT)  \cite{chen2020fedhealth, brophy2021estimation}, drug discovery  \cite{choudhury2019predicting, chen2021fl, xiong2022facing}, and medical image analysis (e.g., X-ray, Ultrasound, CT, MRI, PET, WSIs) \cite{yan2021experiments, feki2021federated, lee2021federated, dou2021federated, florescu2022federated, guo2021multi, stripelis2021scaling, shiri2021federated, andreux2020siloed, adnan2022federated}. Despite the advantages of FL in gigapixel WSIs \cite{andreux2020siloed, wang2023generalizable, baid2022federated, ogier2023federated}, significant challenges remain, particularly in ensuring robustness against considerable heterogeneity among WSIs with a wide range of morphological variations

Data heterogeneity across different centers, characterized by each local model possessing non-identically and independently distributed (non-IID) data, presents a significant challenge in FL \cite{kairouz2021advances}. The divergence between the training and test data distributions can substantially negatively impact model robustness \cite{de2022mitigating}. Several studies have made notable progress in addressing the heterogeneity problem through the following approaches. 
(1) Variance reduction: Recent advancements in this area have focused on balancing the data distribution obtained by centers from other centers to reduce variance \cite{ahn2022communication}. This is achieved by considering various aspects such as parameters, update directions, or feature representations among local models \cite{nguyen2022federated}. 
(2) Global model update: To expedite model convergence and enhance robustness, adaptive \cite{mcmahan2017communication, li2021fedrs, gao2022feddc}  and personalized optimization \cite{hanzely2021personalized} techniques were employed during the model parameter aggregation process. However, it should be noted that these improvements do not specifically address the heterogeneity issue of histopathological images in FL.
By employing these approaches, researchers have made significant strides in mitigating data heterogeneity challenges in FL. Nevertheless, additional studies are required to address the unique characteristics and complexities of histopathological imaging in this context.

Accurate diagnosis of pathological images requires a detailed analysis of features within cancerous regions. In this study, we employed a multiple-instance learning classification model with an attention mechanism to facilitate meticulous examination. From a practical perspective, the model identifies the region of interest of the pathology image and the necessary diagnostic results for physicians, indicating potential applications in pathological imaging. To enhance the final performance of each client model within the FL framework, we aimed to improve the alignment of attention distribution between each client and its corresponding server counterpart models. To achieve this goal, we introduced a novel framework called federated attention consistent learning (FACL). FACL involves two iterative steps: individual client model training and server model fusion. Initially, each client model underwent training using its private dataset. Subsequently, the learned weights from these client models were aggregated to update the central server model. To ensure that each client model accurately reflected the capabilities of the server model, we replicated the server model as an additional replica model for each client in the initial step. Furthermore, to enhance the performance of each client model, we enforced an attentional consistency constraint, ensuring that each client model remained consistent with the server counterpart model when processing the same input images. This enhancement strategy aims to bolster the model's generalization capability, particularly for data samples that have not been encountered previously.

Moreover, although FL retains data locally, there is an inherent risk of private information leakage during parameter transfer to the server \cite{wu2022adaptive, wang2022safeguarding}. To address this concern, we incorporated  differential privacy \cite{dwork2006our} as a protective measure \cite{zhao2022cork}. Notably, most existing advancements in this field have been proposed and validated using a limited number of small-scale image datasets, such as MNIST,  CIFAR-10, CIFAR-100, EMNISTL, and Tiny ImageNet \cite{nguyen2022federated, li2021fedrs, li2021model}. To validate the effectiveness of the proposed model, we curated a large-scale clinical dataset consisting of 19,461 WSIs obtained from multiple centers. Rigorous testing conducted on this dataset, specifically focusing on prostate cancer diagnosis and Gleason grading, demonstrated the clinical grading performance.

Our research makes significant contributions in the following areas:
(1) We propose a novel FL framework named FACL, which ensures privacy preservation and minimizes unnecessary data transmission, aiming to achieve better diagnosis and Gleason grading of prostate cancer based on large-scale and multicenter pathology images.
(2) We introduced an attention consistency algorithm to address the heterogeneity among FL centers. This algorithm improves the generalization ability of the model by maximizing the attention consistency between the local client and server models.
(3) To comprehensively evaluate our approach, we curated a large-scale clinical prostate cancer dataset comprising 19,461 WSIs from multiple medical centers. These diverse data distributions ensure that our evaluation captures the complexities inherent in real-world prostate cancer cases.

% =================================================================================
% ============================== II. Related works  ===============================
% =================================================================================
\section{Related works}
This section presents a comprehensive overview of federated learning, specifically emphasizing its applications in medical image analysis.

\subsection{Federated learning}

Following Google's introduction of the FL framework FedAvg \cite{mcmahan2017communication}, it has developed extensively, particularly in addressing challenging aspects such as non-IID and unbalanced properties.

\begin{table}[!ht]
\centering
\caption{FL implementation scenarios and associated datasets.}
\label{tab_1}
% \begin{tabular}{|p{2.2cm}|p{3.4cm}|p{2.3cm}|} \hline
% \begin{tabular}{p{2.2cm}p{3.4cm}p{2.3cm}} \hline
\begin{tabular}{lll} 
\hline
\textbf{Reference} & \textbf{Application}  & \textbf{Data (Number)} \\ \hline
FedAvg \cite{mcmahan2017communication} & Natural images & \makecell[l]{MNIST \\ CIFAR-10 \\Shakespeare} \\ \hline
FedDif \cite{ahn2022communication}    & Natural images &   \makecell[l]{MNIST \\ FMNIST \\ CIFAR-10} \\ \hline
ComFed \cite{nguyen2022federated} &Natural images & CIFAR-10 \\ \hline
FedRS \cite{li2021fedrs}  &Natural images &  CIFAR-10 \\ \hline
FedDC \cite{gao2022feddc} &Natural images &  CIFAR-10 \\ \hline
FedProx \cite{litian2020federated} &Natural images & \makecell[l]{MNIST \\Sent140 \\ FEMNIST \\ Shakespeare}\\ \hline
FedFAME \cite{malaviya2023fedfame} & Natural images &  \makecell[l]{ CIFAR-10 \\ 20NewsGroups \\ Fashion-MNIST \\ 5AbstractsGroup}  \\ \hline
% FedCL \cite{liu2023fedcl} &   Medical images (X-ray)  &  5908 images  \\ \hline
\cite{dou2021federated}   &   Medical images (CT)     &  10594 images    \\ \hline
\cite{guo2021multi}       &    Medical images (MR)    &  3443 images     \\ \hline
\cite{pati2022federated}  &   \makecell[l]{Medical images \\ (ET, TC, WT)}  & 6314 images  \\ \hline
\cite{saldanha2022swarm} & \makecell[l]{Pathological images}  & \makecell[l]{Epi700 (661) \\ TCGA (632) \\  DACHS (2448) \\ QuASAR (2190)  \\ YCR BCIP (889)}\\ \hline
\cite{lu2022federated} &  \makecell[l]{Pathological images}  & \makecell[l]{RCC (1184) \\  BRCA (2126) \\ CCRCC (511)} \\ \hline
Prop-FFL\cite{hosseini2023proportionally} & \makecell[l]{Pathological images}  & \makecell[l]{Kidney (424) \\ Lung (1217) }\\
\hline
\end{tabular}
\end{table}

To address the non-IID problem, FedDif \cite{ahn2022communication} introduced a diffusion mechanism in which local models diffused in a network, enabling them to learn data distributions beyond their initial experience prior to global aggregation. Following sufficient diffusion iterations, each local model accumulated personalized data distributions resembling the effects observed with training using identically and independently distributed data. ComFed \cite{nguyen2022federated} employs variance reduction techniques on the client side to streamline server aggregation, coupled with global adaptive update techniques for accelerated learning. FedRS \cite{li2021fedrs} proposed the use of "Restricted Softmax" to control the update of missing class weights during the local process. The FedDC \cite{gao2022feddc} introduced a lightweight modification in the local training phase by employing an auxiliary local drift variable to track the gap between the local and global model parameters. As an improvement on FedAvg, FedProx \cite{litian2020federated} incorporated a proximal term into the local objective function. This significantly enhanced the stability of the method, demonstrating its effectiveness. Further, FedFAME \cite{malaviya2023fedfame} introduced a comprehensive framework centered on model-contrastive learning with a primary focus on tackling challenges related to local client data enhancement and non-IID issues, particularly in domains such as texts and graphs. Although these methods have significantly contributed to non-IID problem-solving, it is crucial to acknowledge their primary validation on conventional natural images. Consequently, these approaches cannot be seamlessly applied to the distinctive characteristics of large-scale pathological images prevalent in medical imaging datasets.

FL has garnered widespread attention because of its effectiveness and efficiency in healthcare applications. Dou \textit{et al.} \cite{dou2021federated} demonstrated the feasibility of FL in detecting COVID-19-related CT abnormalities through a multinational study, achieving external validation and demonstrating robust generalization on internal and external datasets. In another study,  Guo \textit{et al.} \cite{guo2021multi}introduced a cross-site MR image reconstruction model that aligned the learned intermediate latent features between different source centers with the latent feature distribution at the target centers. This innovative approach enhanced the performance and adaptability of the model across diverse datasets. Pati \textit{et al.} \cite{pati2022federated}  applied FL to rare disease diagnosis, leveraging data from 71 centers across six continents. The outcome was an automated tumor boundary detector for rare glioblastomas, demonstrating the efficacy and practicality of FL in multicenter settings. These studies collectively underscore the potential of FL in revolutionizing healthcare, emphasizing its adaptability to diverse scenarios, while acknowledging certain constraints in specific applications.

Several studies have demonstrated the potential of FL for analyzing histopathological images. Saldanha \textit{et al.} \cite{saldanha2022swarm} employed the FL model to predict the BRAF mutation status and microsatellite instability in H\&E-stained colorectal cancer pathology sections. Ming \textit{et al.} \cite{lu2022federated} and Seyedeh \textit{et al.} \cite{hosseini2022cluster} utilized FL and differential privacy to underscore their efficacy in safeguarding medical privacy data in the public TCGA dataset. Addressing the issue of fairness among participants, Hosseini \textit{et al.} \cite{hosseini2023proportionally} introduced proportional fair federated learning (Prop-FFL) and substantiated its effectiveness on two pathological datasets. As illustrated in \textbf{Table \ref{tab_1}}, FL finds predominant applications in natural images. However, in the field of medical imaging, its use is more common for non-pathological images. Public datasets are predominantly used for pathological imaging. However, a prevalent limitation of FL research applied to histopathological images is the lack of data diversity. Several studies rely heavily on public datasets, thereby lacking a comprehensive verification of actual clinical data.

% ============================================================
%                         Figure 1
% ============================================================
\begin{figure*}[!hb] 
    \setlength{\abovecaptionskip}{-0cm} 
    \setlength{\belowcaptionskip}{-0.2cm} 
\centering
\includegraphics[width=1\textwidth]{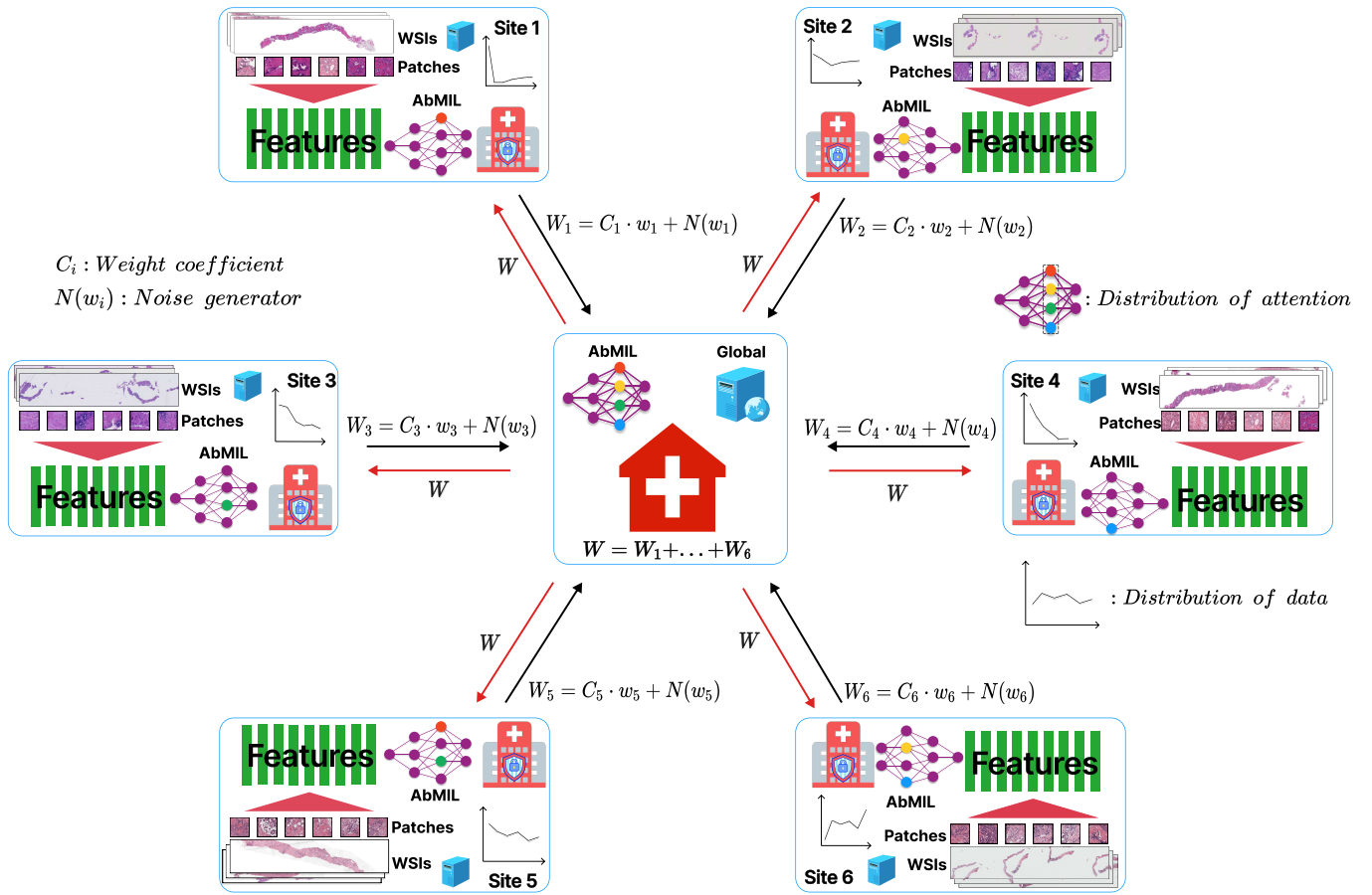}
\caption{Framework of feature extraction and model training process for the FACL model on pathological images}
\label{Fig1}
% \vspace{-0.25cm}
\end{figure*}

% =================================================================================
% =============================== III. Methods  ===================================
% =================================================================================
\section{Methods}

In this section, we provide an overview of the proposed FACL framework illustrated in \textbf{Fig. \ref{Fig1}}. FL retains data locally at individual centers, allowing patch-level segmentation and feature extraction of tissue regions directly from pathological images at the client level. At each client site, the WSIs were cropped into patch images with a size of $224\times 224$ pixels. Subsequently, a Swin transformer, augmented with convolutional operations, was employed to extract discriminative features. These features were then locally preserved as inputs for subsequent model training within the federated learning framework. Each client trains the model by transmitting model parameters to the central server. Subsequently, the server performs weighted averaging by collecting all client model parameters and then distributes the updated parameters to all centers for the next round of training. In addition to these procedures, we enhance dataset protection by incorporating differential privacy. This involves the introduction of random noise during the aggregation of the model parameters at the server.

Initially, each client employs a CTransPath \cite{wang2022transformer}  feature extractor to acquire and store features from their local datasets. Subsequently, within the FACL framework, AttMIL \cite{ilse2018attention}, a weakly supervised model with an attention mechanism, is incorporated to diagnose and grade gigapixel WSIs in prostate cancer while ensuring patient privacy and enhancing model performance trained using individual client data. To address the non-IID challenge, attention-consistency learning was introduced. Despite the imbalanced data distribution among local centers, FACL can rectify these deviations, enabling the model to achieve optimal results. 
In general, different datasets exhibit heterogeneity, and federated learning can mitigate heterogeneity challenges. However, severe heterogeneity was observed in the pathological images. Hence, we propose an attention-consistency mechanism. This mechanism enables each client to obtain attention patterns from other clients after the server aggregates the model updates, thereby allowing each client's model to incorporate diverse attention patterns. This approach enhances the learning capability of individual clients, ultimately leading to more robust performance in the server model.

\subsection{Data preprocessing}

High-resolution histopathological imaging often requires time-consuming analyses and is susceptible to interference from complex backgrounds \cite{wang2023sac}. An integral step in data processing involves cropping the WSIs into smaller sizes suitable for model input. The removal of the background region is crucial because it lacks informative content. Patches ($224\times 224$ pixels) were cropped without overlap in regions with significant tissue presence using the Otsu thresholding method \cite{4310076}. Subsequently, the features of each patch were extracted using the CTransPath \cite{wang2022transformer} feature extractor and aggregated for consolidated storage across the entire image. Notably, unlike the ImageNet-pre-trained model, the CTransPath extractor was pre-trained on multiple histopathology datasets.

\subsection{Weakly-supervised learning on WSIs}

\begin{figure}[!ht] 
    \setlength{\abovecaptionskip}{0cm} 
    \setlength{\belowcaptionskip}{0cm} 
\centering
\includegraphics[width=0.5\textwidth]{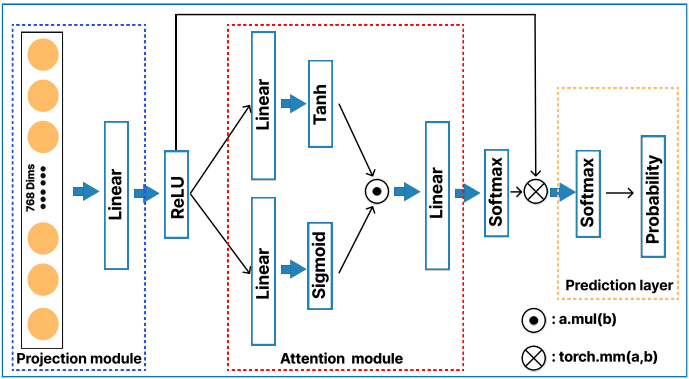}
\caption{Specific implementation details of attention mechanism in FACL}
\label{Fig-AttMIL}
\vspace{-0.25cm}
\end{figure}

At each local center, we employed a weakly supervised model AttMIL, as detailed in \cite{ilse2018attention}. AttMIL integrates an attention mechanism that forms the basis for FL across local centers. \textbf{Fig. \ref{Fig-AttMIL}} depicts the attention mechanism pipeline comprising three modules: $f_{pro}$ for projection, $f_{att}$ for attention, and $f_{pre}$ for prediction. We utilize the gating mechanism in conjunction with $\tanh(\cdot)$ nonlinearity. This mechanism enabled the model to dynamically modulate the importance of different parts of the input sequence when making predictions. In this context, we utilized the feature vector extracted from CTransPath as the input for the model. The attention mechanism facilitates the computation of attention scores for each instance, aiding in the identification of similarities or dissimilarities between instances. This approach, particularly relevant in computational pathology, provides interpretability and reasoning for regions of interest, which would be particularly beneficial for medical professionals in the diagnosis of WSIs.

% ============================================================
%                         Figure 2
% ============================================================

We provide detailed layer-by-layer explanations, beginning with the projection module. This module, comprising contiguous, trainable, and fully connected layers, transforms fixed feature embeddings from a pre-trained encoder into a more compact feature space tailored to histopathological images related to the chosen disease model. The initial fully connected layer $\bm{W_1} \in \mathbb{R}^{512 \times 768}$ compresses 768-dimensional patch-level features $\bm{z}_k$ into a 512-dimensional vector $\bm{h}_k = \bm{W}_1\bm{z}{k}^{\top}$. The resulting set of N patch levels $\bm{h}_k$ in the entire slide image is represented by $\bm{H} \in \mathbb{R}^{N \times 512}$.

If we consider the first two attention matrices, $\bm{U}a$ and $\bm{V}a$, both of dimensions $\mathbb{R}^{256 \times 512}$, as the shared attention backbone for all classes, the attention network can be divided into n parallel attention branches, denoted as $\bm{W}_{a,1}, ..., \bm{W}_{a,n}$, each with dimensions $\mathbb{R}^{1 \times 256}$. Simultaneously, we created $n$ parallel independent classifiers $\bm{W}_{c,1}, ..., \bm{W}_{c,n}$ to assess each class-specific slide-level representation. The attention score for the \textit{k}th patch of the \textit{m}th class, denoted by $a_{k,m}$ and computed using Equation \eqref{eqn1}, contributes to the aggregation of the slide-level feature representation. The resulting slide-level feature representation for the \textit{m}th class is denoted by $h_{slide,m} \in \mathbb{R}^{1 \times 512}$ and is computed using Equation \eqref{eqn2}.
% ============================================================
%                         Equation 1
% ============================================================
\begin{equation}\label{eqn1}
a_{k,m}= \frac{\exp \left\{{W_{a,m}}(\tanh(V_ah_{k}^{\top})\odot \rm{sigm}(\bm{U}_ah_{k}^{\top}) )\right\}}{\sum_{j=1}^{N}\exp \left\{{W_{a,m}}(\tanh(V_ah_{j}^{\top})\odot \rm{sigm}(\bm{U_a}h_{j}^{\top}) )\right\}}
\end{equation}
% ============================================================
%                         Equation 2
% ============================================================
\begin{equation}\label{eqn2}
h_{slide,m} = \sum_{k=1}^{N}a_{k,m}h_k
\end{equation}
The slide-level score $s_{slide,m}$ is derived from the classification layer $\bm{W}_{c,m} \in \mathbb{R}^{1 \times 512}$ using the formula $s_{slide,m} = \bm{W}_{c,m}h_{slide,m}^{\top}$. During the inference, the predicted probability distribution for each class was obtained by applying the SoftMax function to the slide-level prediction scores.

\begin{figure}[!hb] 
    \setlength{\abovecaptionskip}{0cm} 
    \setlength{\belowcaptionskip}{-0.3cm} 
\centering
\includegraphics[width=0.5\textwidth]{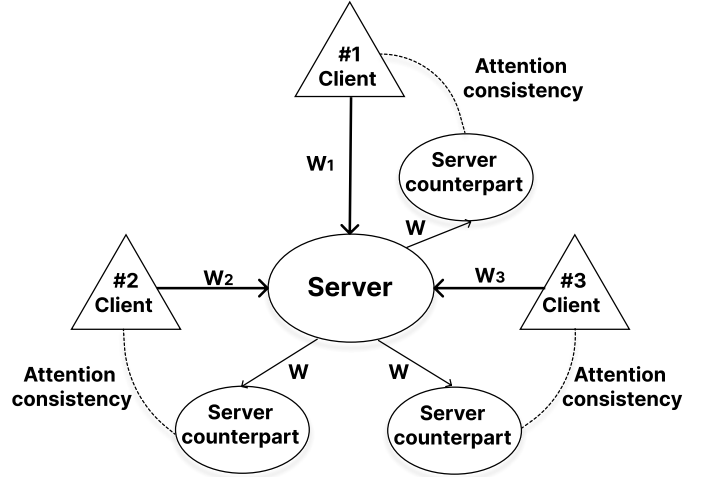}
\caption{Attention consistency learning between server and client models. The process unfolds in two main steps during each iteration of FACL training. First, each client is trained on its private data, generating the weight parameters $W_1$, $W_2$, and $W_3$. Second, these weights are transferred to the server model to obtain average weights $W$. The weight parameter $W$ is then replicated as a copy for each client. When each client and its server counterpart were provided the same input for the next training iteration, we imposed an attention-consistency constraint between the generated attention scores. This approach encouraged each client model to emulate the learning abilities of the server model.}
\label{ACL}
\end{figure}

\subsection{Federated learning on WSIs}\label{sec3.3}
In this section, we elaborate the details of the proposed federated attention-consistent learning method and the differential privacy techniques used to provide additional protection to the model.

\subsubsection{Federated attention consistent learning}

Inspired by contrastive learning, we propose a federated attention-consistency learning method, as shown in \textbf{Fig. \ref{ACL}}. The FACL comprises two repetitive steps: individual client model training and server model fusion. In the first step, each client was trained using private data. In the second step, the weights from the client models were transferred to update the server model. To ensure that each client model possessed capabilities equivalent to those of the server model, we initially duplicated the server model for each client in the first step. Moreover, when the input images were identical, attention consistency constraints were applied between each client ($W_1$, $W_2$, and $W_3$) and its corresponding server counterpart ($W$), thereby enhancing the performance of each client model.
The alignment process for attentional consistency is detailed as follows. In federated learning, the server model collects and synthesizes various attention mechanisms from all the client models. This collective insight is redistributed to each client, allowing them to indirectly benefit from others’ knowledge. Using a copy of the server model, the attention distribution is calculated, which is then considered the baseline or ``ground truth" for the subsequent training phase of the clients. Then, the KL loss between the ``ground truth" attention distribution and the attention distribution computed by the client model is calculated.

The loss of the local model comprises two components. The first part is a standard loss, such as the cross-entropy loss in supervised learning, denoted by $\mathcal{L}_{CE}$ in this study. The second part is the consistent loss, denoted by $\mathcal{L}_{FACL}$, which measures the disparity between the attention scores of the client- and server-side models. Given the predicted probability value of the classification model as ``logits," the label as ``y," and the attention scores of the local model and the server model as $A_{local}$ and $A_s$, respectively, the total loss can be expressed as follows:
% ============================================================
%                         Equation 3
% ============================================================
\begin{equation}
\mathcal{L} = \mathcal{L}_{CE}(logits,y) + \bm{\mu} \mathcal{L}_{FACL}(A_{local},A_{s}),
\end{equation}
% ============================================================
%                         Equation 4
% ============================================================
\begin{equation} 
 \mathcal{L}_{FACL} = KL[P(\bm{X})\parallel Q(\bm{X})]=\sum_{x \in X}^{}[P(x)\log\frac{P(x)}{Q(x)}],
\end{equation}
% ============================================================
%                       Algorithm 1
% ============================================================
\begin{algorithm}[!hb]
    \SetKwData{Left}{left}\SetKwData{This}{this}\SetKwData{Up}{up}
    \SetKwFunction{Union}{Union}\SetKwFunction{FindCompress}{FindCompress}
    \SetKwInOut{Input}{input}\SetKwInOut{Output}{output}
    \caption{Pseudocode of FACL algorithm}
    \label{alg1}
    \KwIn{WSIs Data and weak annotation $ (\bm{X},\bm{Y}) $;}
    \KwOut{Global model $f_{server}$($f_s$);}
    \text{Initialize all model parameters} $\left \{ \bm{W^{0}_g},\bm{W^{0}_1},...,\bm{W^{0}_M} \right \}$ \;
    \small \tcp{\text{Training $K$ rounds}}
    \For{$k =1 $ \KwTo $K$ }{ 
        \small \tcp{ $M$ \text independent centers}
        \For{$i = 1 $ \KwTo $M$}{
            \small \tcp{$N_i$ \text {WSIs}}
            \For{$j = 1 $ \KwTo $N_i$}{ 
                \small \tcp{\text Local model projection}
                $\bm{H}_{i,j} = f_{pro}(\bm{X^{'}_{i,j}}) $ \;
                \small \tcp{\text Server model projection}
                $\bm{H}_{s} = f_{s,pro}(\bm{X^{'}_{i,j}})$ \; 
                \small \tcp{\text Local model attention}
                $\bm{A}_{i,j} = f_{att}(\bm{H}_{i,j})$ \;
                \small \tcp{\text Server model attention}
                $\bm{A}_{s} = f_{att}(\bm{H}_{s})$ \;
                \small \tcp{\text Feature representation of the WSI}
                $\bm{h}_{bag} = \rm{W\text{-}Ave} (\bm{A_{i,j}},\bm{H}_{i,j})$ \;
                \small \tcp{\text Output predicted value}
                $s_{i,j} = f_{pre}(\bm{h}_{bag_{i,j}}) $ \;
                \small \tcp{\text Calculate attention consistency}
                $\mathcal{L}_{FACL}(\bm{A_{i,j}},\bm{A}_{s}) = KL[\bm{A_{i,j}} \parallel \bm{A}_{s}] $ \; 
                \small \tcp{\text Calculate loss function}
                $\mathcal{L} = \mathcal{L}_{CE}({s}_{i,j}, Y_{i,j}) + \mu \mathcal{L}_{FACL}(\bm{A_{i,j}},\bm{A}_{s}) $ \;
                % $\bm{W}^{(k)}_i  \leftarrow {\rm Opt}_i(\mathcal{L}, \bm{W}^{(k)}_i) $ \;
                \small \tcp{\text Optimize model parameters}
                $\bm{W}_i  \leftarrow {\rm Opt}_i(\mathcal{L}, \bm{W}_i) $ \;
            }
       } 
        \small \tcp{\text Update server parameters}
        \small \tcp{$\gamma_i$:\text{Weight coefficient}}
        \small \tcp{$N_g(\bm{W})$:\text{Noise generater}} 
        $\bm{W}_{s} \leftarrow \sum_{i}^{}\gamma_i(\bm{W}_i) + N_g(\bm{W}_i)$ \;
        \small \tcp{\text Update local model parameters}
        \For{$i = 1 $ \KwTo $M$}{
            % $ (\bm{W}^{(k)}_i) \leftarrow \bm{W}^{(k)}_{s}$ \;
            $ \bm{W}_i \leftarrow \bm{W}_{s}$ \;
        } 
    }
    \text{return} $f_s$
\end{algorithm}
Here, $\bm{\mu}$ represents the hyperparameter controlling the weight of attention-consistent loss. $\mathcal{L}_{FACL}$ denotes the Kullback-Leibler divergence \cite{joyce2011kullback}, which quantifies the similarity between two probability distributions.

\subsubsection{FACL with differential privacy}

By embracing the principles of data differential privacy, we seamlessly integrate this approach into the FL server, specifically during the phase in which the server receives parameter updates from individual client models. The crux of differential privacy involves introducing noise into model weights before transmitting them to a centralized server. To achieve this, we utilized a noise generator denoted by \(N_g(\cdot)\), generating Gaussian random noise \(\bm{N}\sim(0, \bm{z}^2 \eta^2)\), where \(\bm{z}\) indicates the noise level and \(\eta\) represents the standard deviation of the weight of the neural network. The FedAvg model with added noise and the FACL model are denoted as FedAvg-\(N\) and FACL-\(N\), respectively. The noise is meticulously adjusted to prevent a significant deterioration of the learned weights or overall model performance. Achieving an optimal balance entails introducing adequate noise to safeguard privacy while allowing the model to extract valuable information \cite{adnan2022federated, ziller2021medical}. In FL, noisy weight updates from multiple centers are consolidated on the server. Although the introduction of noise into the weights may have a marginal impact on the learned weights, FL was designed to minimize this effect.

By strategically managing added noise and aggregating updates from multiple centers, the FACL effectively maintains commendable model performance while ensuring privacy. In this investigation, we aimed to strike a delicate balance between privacy preservation and model efficacy. We conducted experiments by introducing noise of varying magnitudes and discovered that a noise factor of \(z=0.1\) provides optimal protection without compromising the overall performance of the model. The culmination of our efforts is encapsulated in the proposed FACL algorithm, which integrates attention-consistent learning and differential privacy, detailed in \textbf{Algorithm \ref{alg1}}. 
In each iteration, the server disseminates the global model to all the participating parties for localized training. Subsequently, the updated local models from each party contribute to refining the global model using a weighted average. The number of FL training rounds is denoted by \(K\). Please assume that the WSIs data and its weakly supervised labels are distributed over \(M\) independent centers:
$(\bm{X}, \bm{Y}) = \left\{ \left\{ (\bm{X}_{1,j}, Y_{1,j}) \right\},...,\left\{ (\bm{X}_{M,j}, Y_{M,j}) \right\} \right\}$, where $\left\{ (\bm{X}_{i,j}, Y_{i,j}) \right\} = \left \{ (\bm{X}_{i,1}, Y_{i,1}),...,(\bm{X}_{i, N_i}, Y_{i, N_i})  \right \}$ denotes the $N_i$ pair of WSIs data and the corresponding labels for the $i$th client center, $\bm{X}_{i,j}$ is a single WSI.  After data preprocessing and feature extraction, the data are represented as $(\bm{X}', \bm{Y}) = \left \{  \left \{(\bm{X}'_{1,j}, Y_{1,j}) \right \},..., \left \{(\bm{X}'_{M,j}, Y_{M,j})\right \}  \right \}$. The server-side model is $f_s$ and the local client's neural network model $\bm{f}_{center} = \left \{\bm{f}_1,..., \bm{f}_M \right \}$. Each model $\bm{f}_i$ contains a projection module $f_{i, pro}$, attention module \(f_{i, att}\), and prediction layer \(f_{i, pre}\). W-Ave: The attention scores predicted by all patches are used as weights to average the feature embeddings of all patches. The parameters of the local model are denoted by \(\bm{W}_1,...,\bm{W}_M\) and those of the global model are \(\bm{W}_g\). \({\rm Opt}_i\) is an optimizer that updates the model parameters.
% =================================================================================
% ========================== IV. Experiments and results  =========================
% =================================================================================

\section{Experiments and results}

This section introduces the datasets used for prostate cancer diagnosis and Gleason grading, providing details on the data distribution and experimental setup. We assess the effectiveness of FL in both diagnostic and Gleason grading tasks and compare the performance differences between the FedAvg and single-center models. Additionally, we utilize FedAvg and FedAvg-$N$ as baseline comparisons for FACL and compare them with the results of the FACL and FACL-\(N\) models, respectively. To enhance interpretability, a heat map is employed to visualize the cancerous regions identified by the model.

% ============================================================
%                             Table II & III
% ============================================================
%  Hebei-1, Hebei-2, Nanchang, DiagSet-B-1, DiagSet-B-2, PANDA-1, PANDA-2
\begin{table}[!ht] 
		\setlength{\abovecaptionskip}{0cm}    % # Adjust the spacing
		\setlength{\belowcaptionskip}{0cm}
\caption{Data distribution used for cancer diagnosis (x\% denotes the proportion of positive slides)}
\label{tab_2}
\centering
\setlength{\tabcolsep}{5pt}
\begin{tabular}{lccc}
\hline
% \toprule
\textbf{Center} & \textbf{WSIs No.}  & \textbf{Train No.} &  \textbf{Val No.} \\ \hline
  Hebei-1      & 1201 (21\%)  &   960  & 241 \\
  Hebei-2      & \enspace 844 (73\%) &  675  &  169   \\
  Nanchang     & \enspace 983 (90\%) &  786  &  197   \\
  DiagSet-B-1                   & 2313 (45\%) &  1850  &  463   \\ 
  DiagSet-B-2                   & 2313 (45\%) &  1850  &  463  \\ 
  PANDA-1                       & 5453 (65\%) &  4362  & 1091  \\ 
  PANDA-2                       & 5159 (81\%) &   4127 &   1032  \\ \hline
\end{tabular}
\end{table}
% Hebei-1, Hebei-2, PANDA-1-1, PANDA-1-2, PANDA-2-1, PANDA-2-2  
\begin{table*}[!hb]
		\setlength{\abovecaptionskip}{0cm}   % # Adjust the spacing   
		\setlength{\belowcaptionskip}{0cm}
\caption{\label{tab_3}Data distribution used for cancer diagnosis (The x\% represents the proportion of each class in the total data volume)}
\centering
\begin{tabular}{cccccccc}
\hline
\textbf{Center} & \textbf{WSIs No.} & \textbf{ISUP-0} & \textbf{ISUP-1} & \textbf{ISUP-2} &\textbf{ISUP-3} & \textbf{ISUP-4} & \textbf{ISUP-5} \\
\hline
Hebei-1 & 1201  & 952 (79.3\%)  & \enspace 18 (1.5\%)  & \enspace 17 (1.4\%) & \enspace 42 (3.5\%)  & \enspace 72 (6.0\%)  & 100 (8.3\%) \\
Hebei-2  & \enspace 844  & 220 (26.1\%)  & 169  (20.0\%) & \enspace 88 (10.4\%) & 108 (12.8\%) & 125 (14.8\%)& 134 (15.9\%) \\
PANDA-1-1 & 2727 & 962 (35.3\%)  & 906 (33.2\%) & 334 (12.2\%) & 159 (5.8\%) & 241 (8.8\%) & 125 (4.6\%) \\
PANDA-1-2 & 2727 & 962 (35.3\%) & 907 (33.3\%)  & 334 (12.2\%)  & 158 (5.8\%) & 240 (8.8\%) & 126 (4.6\%) \\
PANDA-2-1 & 2580 & 484 (18.8\%) & 426 (16.5\%) & 338 (13.1\%) & 462 (17.9\%) & 384 (14.9\%) & 486 (18.8\%) \\ 
PANDA-2-2 & 2579 & 483 (18.7\%) & 426 (16.5\%) & 337 (13.1\%) & 462 (17.9\%) & 384 (14.9\%) & 487 (18.9\%) \\  \hline 
\end{tabular}
\end{table*}

\subsection{Datasets}

In this study, we investigated the diagnosis of prostate cancer using a two-level classification approach to differentiate between benign and malignant conditions. Furthermore, we examined the Gleason grading of prostate cancer by employing a six-level classification system based on the International Society of Urological Pathology (ISUP \cite{bulten2022artificial}) categories 0, 1, 2, 3, 4, and 5. A series of preprocessing steps was implemented on the datasets to demonstrate the efficacy of the FL model. These steps aim to enhance the heterogeneity of the datasets by splitting those with a substantial number of samples into multiple independent centers, thus expanding the number of client centers within the FL model. The distribution of the datasets utilized in the cancer diagnosis task is presented in \textbf{Table \ref{tab_2}}. The DiagSet-B and PANDA datasets were split into two separate centers, DiagSet-B-1, DiagSet-B-2, PANDA-1, and PANDA-2, with the proportion of positive data ranging from 21\% to 90\%. Similarly, the dataset distribution for Gleason grading is illustrated in \textbf{Table \ref{tab_3}}, the PANDA datasets have been split into four separate centers, namely PANDA-1-1, PANDA-1-2, PANDA-2-1, and PANDA-2-2, where the categories of ISUP 0-5 exhibit a notable imbalance. According to the definition of federated learning classification, our study uses horizontal federated learning \cite{yang2019federated}.

For the prostate cancer diagnosis task, we curated datasets from three hospitals (Hebei-1, {Hebei-2, and Nanchang) and two public sources (DiagSet-B and PANDA) for training. To evaluate the effectiveness of our approach, we tested a private hospital dataset (QHD) and a public dataset (DiagSet-A). Hebei-1 and Hebei-2 represent hospital datasets from the Hebei Province, China, while Nanchang denotes a hospital dataset from Nanchang, China. The QHD dataset, sourced from a hospital in Qinhuangdao, China, comprised 765 pathological images, 433 of which were positive. DiagSet-A, a subset of the DiagSet data, comprised 430 pathological images, of which 227 were positive.

To assess the accuracy of prostate cancer Gleason grading, we employed two private datasets (Hebei-1 and Hebei-2) and one public dataset (PANDA) for training purposes. In the evaluation phase, a private hospital dataset (Nanchang) was utilized. This allowed us to evaluate the performance and reliability of the proposed approach. 

\subsection{Experimental setups and evaluation metrics}

We utilized the loss functions and algorithmic models outlined in Section \ref{sec3.3} for each task. As an illustration, we conducted a simulated FL experiment employing 8 Tesla V100 GPUs, with each card representing an independent center. In this experiment, we employed a learning rate of 0.0002, an Adam optimizer, and a batch size of 1. To ensure optimal results, the model was trained for over 60 epochs. Additionally, we implemented an early stopping strategy, halting training when the server model failed to demonstrate improvement after 20 consecutive epochs while ensuring that a minimum of 40 epochs were completed. To promote convergence toward optimal outcomes, we calculated the validation metrics for each epoch and selected the model with the highest performance in the validation set for the final evaluation of the test set.

Our experiments encompassed three distinct settings: (1) local learning: Each center independently trained its data; (2) global learning: aggregating datasets from all centers, along with training a multicenter centralized model; (3) federated learning: Training on data from all centers without sharing datasets, focusing solely on exchanging model parameters. To comprehensively validate the effectiveness of attention-consistency learning, we introduced a noisy model, FedAvg-N, as another baseline. Noise introduces significant interference into the model, and we aim to ascertain whether attention consistency still holds under noisy conditions. By employing these strategies, we aim to ensure a comprehensive evaluation of our model across different learning scenarios.
% ============================================================
%                         Table 4
% ============================================================
\begin{table}[!hb]
\renewcommand\arraystretch{1.2} 
		\setlength{\abovecaptionskip}{0cm}    %  # Adjust the spacing   
		\setlength{\belowcaptionskip}{0cm}
\caption{\label{tab_4} The proportion of positive data across four centers} 
{
\begin{tabular}{ccccc}
\hline
\bm{$\alpha$} & \textbf{DiagSet-B-1} & \textbf{DiagSet-B-2} & \textbf{PANDA-1} & \textbf{PANDA-2} \\ \hline
 0.05 & 50/1000 & 50/1000  &75/1500  & 1425/1500         \\ 
 0.1  & 100/1000  & 100/1000  & 150/1500  & 1350/1500   \\ 
 0.3  & 300/1000 & 300/1000 & 450/1500  & 1050/1500    \\ 
 0.5  & 500/1000 & 500/1000 &  750/1500 & 750/1500    \\ \hline
\end{tabular}
}
\end{table}

The distribution of data splits for the diagnostic task is presented in \textbf{Table \ref{tab_4}}. To assess the impact of data imbalance on the model's performance, we partitioned the dataset from four centers (DiagSet-B-1, DiagSet-B-2, PANDA-1, PANDA-2), which contains a substantial number of samples, into positive proportions denoted as  $\alpha$ (specifically, $\alpha \in \{0.05, 0.1, 0.3, 0.5\}$). Notably, we focused on conducting this verification solely within the realm of binary classification, primarily because of significant variations in the quantity and category of the datasets. Therefore, we did not conduct a similar distribution experiment in the context of multiclass classification.
Consequently, the FL simulation experiment involved seven centers. Each center's dataset was randomly divided into training and validation sets at a ratio of 80\% and 20\%, respectively. Stratified sampling was employed to ensure a consistent class balance between each center's training and validation sets. Furthermore, we used datasets from two independent centers to evaluate the robustness and generalization of the model to previously unseen datasets for testing. For the Gleason grading task, we implemented an FL model with six centers. Additionally, we employed an independent dataset obtained from a private institution for testing. On diagnostic and Gleason grading tasks, we evaluated the proposed FACL model using a variety of classification metrics, including area under the curve (AUC), F1 score, ACC, Recall, and Kappa score.

% ============================================================
%             Table V. Validation-1 (Diagnostic model)
% ============================================================
\begin{table}[!ht]
		\setlength{\abovecaptionskip}{0cm}   % Adjust the spacing   
		\setlength{\belowcaptionskip}{0cm}
\renewcommand\arraystretch{1}
\caption{\label{tab_5}The performance of the diagnostic model on the validation set was reported as the mean of five-fold cross-validation results}
\centering
\begin{tabular}{llllll}
\toprule 
\textbf{$\alpha$} &  \makecell[l]{\textbf{Training}  \\ \textbf{settings}}  & \textbf{AUC}   & \textbf{F1}  & \textbf{ACC}   & \textbf{Recall}   \\ 
\hline 
 & Hebei-1     & 0.9487   & 0.8481   & 0.8325  & 0.7450     \\ 
/ & Hebei-2    & 0.9422   & 0.8973   & 0.8698  & 0.8981     \\ 
 & Nanchang    & 0.9410   & 0.8685   & 0.8174  & 0.9303     \\    
 \hline
 & DiagSet-B-1         & 0.9222  & 0.7917  & 0.7830   & 0.6633   \\ 
 & DiagSet-B-2         & 0.9385  & 0.7679  & 0.7620   & 0.6264   \\ 
 & PANDA-1             & 0.9510  & 0.8999  & 0.8821   & 0.8426   \\ 
 & PANDA-2             & 0.9312  & 0.8439  & 0.7716   & 0.9622   \\ 
 & Avg. (7 centers)    & 0.9393  & 0.8453  & 0.8169   & 0.8097   \\
 0.05   & Centralized  & 0.9680  & 0.9291  & 0.9129   & 0.9022   \\ 
 & FedAvg              & 0.9620  & 0.9247  & 0.9083   & 0.8887   \\ 
 & FedAvg-$N$       & 0.9633  & 0.9248  & 0.9089   & 0.8855   \\ 
 & \textbf{FACL}     & 0.9626  & 0.9260  & 0.9096   & 0.8936   \\ 
 & \textbf{FACL-$N$}  & \textbf{0.9635}   & 0.9256   & 0.9099  & 0.8859   \\ 
\hline
 & DiagSet-B-1           & 0.9402 & 0.8419   & 0.8273 & 0.7326   \\ 
 & DiagSet-B-2           & 0.9513 & 0.8177  & 0.8054 & 0.6965   \\ 
 & PANDA-1          & 0.9487 & 0.9012   & 0.8838 & 0.8387   \\ 
 & PANDA-2           & 0.9514 & 0.8830   & 0.8390  & 0.9555   \\ 
 & Avg. (7 centers)    & 0.9462   & 0.8654 & 0.8393  & 0.8281   \\
0.1  & Centralized        & 0.9762 & 0.9392   & 0.9251   & 0.9152   \\ 
& FedAvg               & 0.9691 & 0.9326   & 0.9179  & 0.8963   \\ 
 & FedAvg-$N$         & 0.9708 & 0.9304   & 0.9160  & 0.8876   \\ 
 & \textbf{FACL}     & 0.9701 & 0.9343   & 0.9201  & 0.8976   \\ 
 & \textbf{FACL-$N$} & \textbf{0.9715} & 0.9345   & 0.9204  & 0.8970\\ 
\hline
 & DiagSet-B-1         & 0.9595 & 0.8583  & 0.8422 & 0.7566   \\ 
 & DiagSet-B-2         & 0.9436 & 0.8415  & 0.8269 & 0.7324  \\ 
 & PANDA-1             & 0.9570  & 0.9176   & 0.8946 & 0.9257  \\ 
 & PANDA-2           & 0.9566 & 0.9143   & 0.8898 & 0.9257   \\ 
 & Avg. (7 centers)    & 0.9498   & 0.8780 & 0.8533  & 0.8448   \\
 0.3 & Centralized        & 0.9842 & 0.9511 & 0.9397 & 0.9265   \\ 
 & FedAvg                 & 0.9725  & 0.9382   & 0.9245 & 0.9059   \\ 
 & FedAvg-$N$       & 0.9722 & 0.9341   & 0.9198  & 0.8977  \\ 
 & \textbf{FACL}     & 0.9731  & 0.9420    & 0.9291 & 0.9099   \\ 
 & \textbf{FACL-$N$} & \textbf{0.9743} & 0.9447   & 0.9318 & 0.9202   \\ 
\hline
 & DiagSet-B-1          & 0.9522  & 0.8801   & 0.8634 & 0.7971  \\ 
 & DiagSet-B-2           & 0.9570  & 0.8934   & 0.8768 & 0.8202   \\ 
 & PANDA-1            & 0.9572 & 0.8977   & 0.8596 & 0.9440   \\ 
 & PANDA-2          & 0.9516  & 0.9144   & 0.8947 & 0.8873 \\ 
 & Avg. (7 centers)    & 0.9499   & 0.8856 & 0.8592  & 0.8603   \\
 0.5 & Centralized        & 0.9837 & 0.9540    & 0.9428 & 0.9371   \\
 & FedAvg               & 0.9671  & 0.9386   & 0.9242 & 0.9148   \\ 
 & FedAvg-$N$         & 0.9682 & 0.9357   & 0.9210 & 0.9087   \\ 
 & \textbf{FACL}      & 0.9716 & 0.9435   & 0.9303 & 0.9201   \\ 
 & \textbf{FACL-$N$}  & \textbf{0.9718} & 0.9412   & 0.9269 & 0.9238   \\ 
\bottomrule
\end{tabular}
\end{table}

\subsection{Results of diagnosis (Binary classification)}

In the diagnostic task, we assessed models trained on datasets with varying degrees of heterogeneity, as indicated by the proportion factor $\bm{\alpha} \in \left \{0.05, 0.1, 0.3, 0.5 \right \}$ in \textbf{Table \ref{tab_4}}. Our experiments aimed to validate the performance of each local center, FedAvg, FACL, and \textit{Centralized} model across seven internal datasets and two external datasets (DiagSet-A and QHD).

\subsubsection{Internal datasets}
% ============================================================
% Table VI Test-1.1(Diagnostic model)  FL  test-1.1 (DiagSet-A) set and  test-1.2 (QHD) 
% ============================================================
\begin{table*}[!hb]
\renewcommand\arraystretch{1} 
		\setlength{\abovecaptionskip}{0cm}    % Adjust the spacing   
		\setlength{\belowcaptionskip}{0cm}
\centering
\caption{\label{tab_6} Comparison results between FedAvg and FACL on DiagSet-A and QHD sets}
\begin{tabular}{llllllllll} 
\toprule       
           &             & \multicolumn{4}{c}{\textbf{DiagSet-A}}   & \multicolumn{4}{c}{\textbf{QHD}}  \\ 
\textbf{$\alpha$}  & \textbf{FL models}  &\textbf{AUC}   & \textbf{F1}  & \textbf{ACC}   & \textbf{Recall}   &\textbf{AUC}  & \textbf{F1}  & \textbf{ACC}   & \textbf{Recall}    \\
\hline
\multirow{4}{*}{0.05}   & FedAvg   & 0.9635 & 0.9079 & 0.8969 & 0.9559    & 0.9795  & 0.9439  & 0.9346 & 0.9722  \\
& FedAvg-$N$     & 0.9632 & 0.9156 & 0.9063 & 0.9559     & 0.9804 & 0.9491 & 0.9411 & 0.9699 \\
& \textbf{FACL}       & 0.9666 & 0.9188 & 0.9110 & 0.9471   & \textbf{0.9825} & 0.9461  & 0.9372  & 0.9722 \\
& \textbf{FACL-$N$} & \textbf{0.9674} & 0.9148 & 0.9063 & 0.9471   & 0.9824  & 0.9379  & 0.9267 & 0.9769\\
\hline 
\multirow{4}{*}{0.1} & FedAvg   & 0.9648 & 0.9204 & 0.9133 & 0.9427    & 0.9775     & 0.9426      & 0.9333      & 0.9676     \\
& FedAvg-$N$   & 0.9651 & 0.9135 & 0.9063 & 0.9251     &\textbf{0.9782}     & 0.9553      & 0.9490      & 0.9630  \\
&  \textbf{FACL}         & 0.9678   & 0.9091    & 0.9016  & 0.9251      & 0.9758   & 0.9486      & 0.9411      & 0.9607                \\
& \textbf{FACL-$N$} & \textbf{0.9687} & 0.9164    & 0.9086   & 0.9427    & 0.9766 & 0.9466  & 0.9385 & 0.9630          \\
\hline
\multirow{4}{*}{0.3} & FedAvg & 0.9651 & 0.8972 & 0.8852 & 0.9427    & 0.9788     & 0.9523       & 0.9451      & 0.9699  \\
& FedAvg-$N$   & 0.9664     & 0.9121    & 0.9016     & 0.9603     & 0.9789     & 0.9490       & 0.9411      & 0.9676                 \\
& \textbf{FACL} & 0.9710   & 0.9048    & 0.8946     & 0.9427    & \textbf{0.9801}  & 0.9500       & 0.9424      & 0.9653             \\
& \textbf{FACL-$N$} & \textbf{0.9741} & 0.9063  & 0.8969    & 0.9383   & 0.9792       & 0.9239       & 0.9098      & 0.9676        \\
\hline 
\multirow{4}{*}{0.5} & FedAvg     & 0.9647   & 0.8957  & 0.8805  & 0.9647     &   0.9765        & 0.9522      & 0.9451    & 0.9676           \\
& FedAvg-$N$   & 0.9661   & 0.9012    & 0.8875  & 0.9647     & 0.9783        & 0.9478      & 0.9398    & 0.9653               \\
& \textbf{FACL}     & 0.9718         & 0.8934   & 0.8782    & 0.9603    & \textbf{0.9796} & 0.9457    & 0.9372   & 0.9653 \\
& \textbf{FACL-$N$} & \textbf{0.9725} & 0.8915    & 0.8735     & 0.9779  & 0.9762     & 0.9170    & 0.9006   & 0.9699   \\
\bottomrule
\end{tabular}
\end{table*}
The experimental results for the diagnosis task on the validation set are presented in \textbf{Table \ref{tab_5}}, demonstrating metrics such as AUC, F1, ACC, and Recall. As $\alpha$ increases, the overall performance of the local center model improves due to the different proportions of categories in the diagnostic task. When $\alpha$ is 0.05, the average ACC of the models from seven centers (Hebei-1, Hebei-2, Nanchang, DiagSet-B-1, DiagSet-B-2, PANDA-1, PANDA-2) is only 0.8169, with an average AUC of 0.9393. However, when $\alpha$ is 0.5, the average ACC of these models (Hebei-1, Hebei-2, Nanchang, DiagSet-B-1, DiagSet-B-2, PANDA-1, PANDA-2) reaches 0.8592, accompanied by an average AUC of 0.9499. 

We observed superior performance in the FL models compared to the local center models. Specifically, the FACL model demonstrated a notable enhancement in AUC, ranging from 0.02 to 0.04 when compared to individual centers. The deviation in the results between the FACL model and the \textit{Centralized} model is minimal, with less than a 0.01 difference in AUC. This outcome suggests that the FACL model achieved an optimal value in line with federated learning theory, closely aligning with the performance of the \textit{Centralized} model.

\subsubsection{External datasets (DiagSet-A and QHD)}

WSIs can vary widely in morphological appearance due to differences in institutional standards and protocols for tissue processing, slide preparation, and digitization of pathological images. Therefore, it is important to generalize the application of the model to actual clinical data. We validate the robustness of the model using two external datasets. Our findings indicate that models trained using FL on multicenter data significantly improved the generalization performance of the models compared to models trained on single-center data. 

The FedAvg and FACL models for the test sets are listed in \textbf{Table \ref{tab_6}}. The FACL model demonstrated superior AUC performance compared with the FedAvg model for both the DiagSet-A and QHD datasets. Similarly, the FACL-$N$ model outperformed the FedAvg-$N$ model in terms of AUC performance. 
For the DiagSet-A dataset, FedAvg-$N$ achieved optimal results, demonstrating its effectiveness. In addition, on the QHD dataset, the FACL achieved optimal results without added noise. However, the results remained highly satisfactory even when noise was introduced into the FACL-$N$ model. The comparison results between the single-center and \textit{Centralized} models can be found in the tables \ref{tab_s1} and \ref{tab_s2} in the appendix. 

% ============================================================
%           Table VII  Validation-2 (Gleason grading)
% ============================================================
\begin{table}[!ht]
\renewcommand\arraystretch{1}	
		\setlength{\abovecaptionskip}{0cm}    % Adjust the spacing   
		\setlength{\belowcaptionskip}{0cm}	
\caption{\label{tab_7}The performance of the Gleason grading model on the validation set was reported as the mean of a five-fold cross-validation}
\centering
\begin{tabular}{lllll}
\toprule
\textbf{Training settings}  & \textbf{Kappa}  & \textbf{AUC}  & \textbf{F1}  &  \textbf{Recall}    \\ \hline
 Hebei-1               & 0.6491   & 0.7679 & 0.3699  & 0.4589    \\
 Hebei-2               & 0.6977     & 0.7989 & 0.4787  & 0.5040   \\
 PANDA-1-1               & 0.8094    & 0.8866 & 0.6176  & 0.6367   \\
 PANDA-1-2               & 0.7531    & 0.8716 & 0.5476  & 0.5772   \\
 PANDA-2-1               & 0.7633   & 0.8748 & 0.6007  & 0.6218   \\
 PANDA-2-2               & 0.7552   & 0.8684 & 0.5794  & 0.6122   \\
 Avg. (6 centers)    & 0.7379   & 0.8447 & 0.5323  & 0.5684  \\
 Centralized             & 0.8779    & 0.9373  & 0.7490   & 0.7515 \\
\toprule
FedAvg                                       & 0.8494  & 0.9212 & 0.7158  & 0.7197   \\
FedAvg-$N$       & 0.8444  & 0.9136 & 0.6877  & 0.6951   \\
\textbf{FACL}      & 0.8535  & 0.9214 & 0.7201  & 0.7251    \\
\textbf{FACL-$N$}  & 0.8463  & 0.9126 & 0.6892  & 0.6969    \\
\bottomrule
\end{tabular}
\end{table}

% ============================================================
%                 Table 8   Test-2 (Gleason grading)
% ============================================================
\begin{table}[!htbp]
		\setlength{\abovecaptionskip}{0cm}   % # Adjust the spacing   
		\setlength{\belowcaptionskip}{0cm}
\caption{\label{tab_8} The Gleason grading model's performance on Nanchang was reported as the five-fold mean}
\centering
\begin{tabular}{lllll}
\toprule
\textbf{Training settings}  & \textbf{Kappa}  & \textbf{AUC}  & \textbf{F1}  & \textbf{Recall}   \\
\hline
 Hebei-1               & 0.5494           & 0.7540 & 0.3107 & 0.3581   \\
 Hebei-2               & 0.6142           & 0.7862 & 0.3652 & 0.3977   \\
 PANDA-1-1             & 0.7052           & 0.8259 & 0.4525 & 0.4954   \\
 PANDA-1-2             & 0.6346           & 0.8423 & 0.3242 & 0.3794   \\
 PANDA-2-1             & 0.6649           & 0.8186 & 0.3648 & 0.4292   \\
 PANDA-2-2             & 0.6936           & 0.8307 & 0.3326 & 0.4374   \\
 Avg. (6 centers)      & 0.6437   & 0.8096 & 0.3583 & 0.4162   \\
 FedAvg                & 0.7247   & 0.8435 & 0.4552 & 0.4954   \\
 Centralized           & 0.7171   & 0.8365 & 0.4301 & 0.4638   \\
% \hline
\bottomrule
\end{tabular}
\end{table}

\subsection{Results of Gleason grading (Multiple classifications)}

Given the scarcity of ISUP-annotated datasets, we assessed the performance of the local center models, FedAvg model, FACL model, and centralized model on five internal datasets and an external dataset (Nanchang).

\subsubsection{Internal datasets}

% ============================================================
%                       Figure 4  Heatmap
% ============================================================
\begin{figure*}[!hb]
    \setlength{\abovecaptionskip}{0cm} 
    \setlength{\belowcaptionskip}{-0.2cm} 
\centering 
\includegraphics[width=1\textwidth]{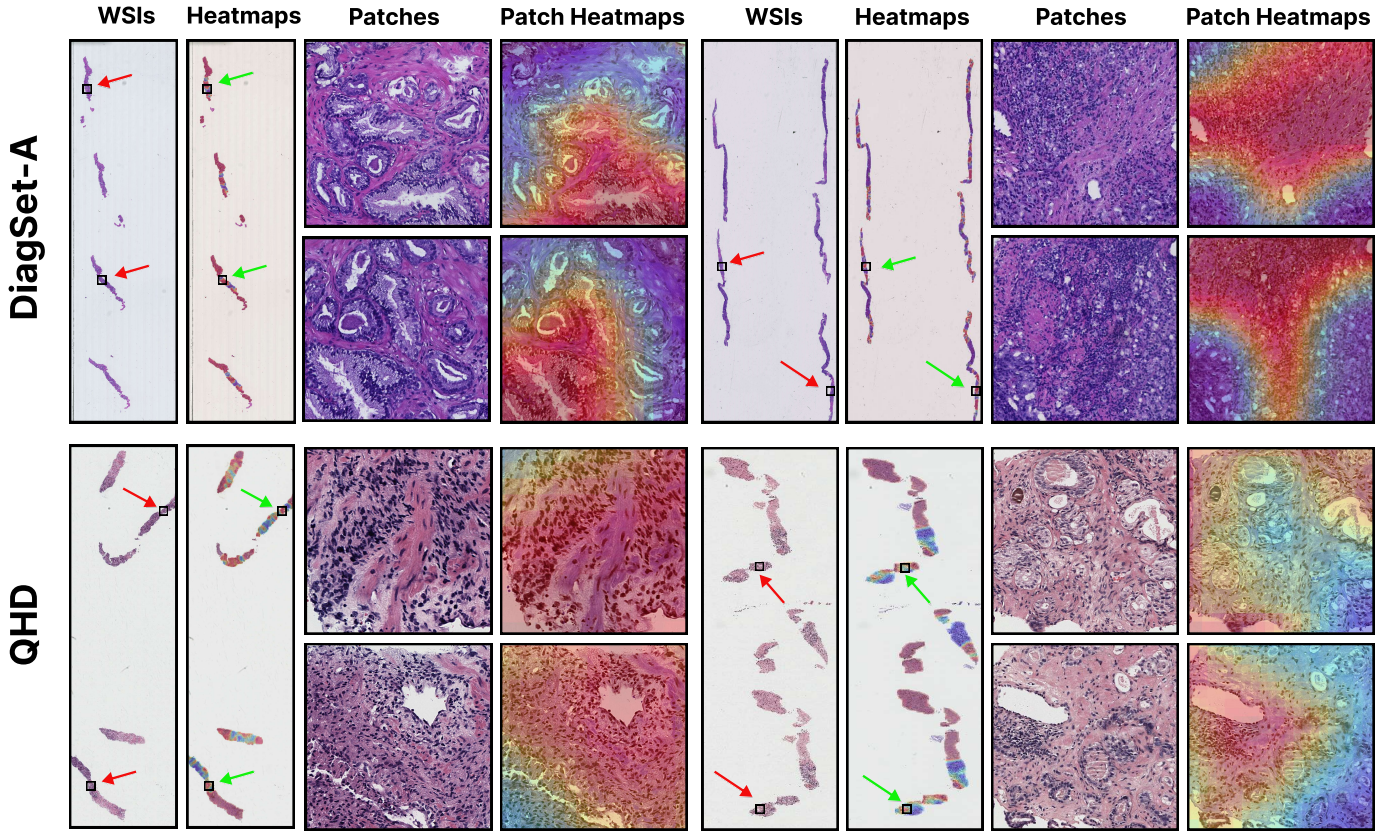} 
\caption{Interpretability and visualization of FACL. Randomized selections of WSIs from external datasets, DiagSet-A and QHD, are arranged in the first and second rows, respectively. The sequence progresses from left to right, showcasing the complete WSI, followed by its heatmap, a close-up of a local patch, and finally the heatmap of the patch. The zoomed-in view of the local patch image indicates our model's precise identification and representation of cancerous regions.}
\label{Fig5}
\end{figure*}

The validation results for the Gleason scoring task are listed in \textbf{Table \ref{tab_7}}. Compared to models trained on single-center data, the FACL model exhibited significant improvements in the Kappa score and AUC. The average Kappa across the six centers (Hebei-1, Hebe-2, PANDA-1-1, PANDA-1-2, PANDA-2-1, and PANDA-2-2) was 0.7379, whereas FACL achieved a Kappa of 0.8463. This highlights the effectiveness of federated learning for prostate cancer diagnosis across multiple categories. Notably, the proposed FACL model consistently outperformed the FedAvg model in terms of the Kappa score, regardless of the addition of noise ($N$). The Kappa scores for the FACL surpassed those for the FedAvg, and the FACL-$N$ outperformed the FedAvg-$N$. This underscores the efficacy of the proposed attention-consistent learning method.

\subsubsection{External datasets (Nanchang)}
% ============================================================
%                 Table 9   Test-2 (Gleason grading)
% ============================================================
\begin{table}[!ht]
		\setlength{\abovecaptionskip}{0cm}   % # Adjust the spacing   
		\setlength{\belowcaptionskip}{0cm}
\caption{\label{tab_9}Comparison results between FedAvg and FACL on the Nanchang set}
\centering
\begin{tabular}{lllll}
\toprule
\textbf{FL models}  & \textbf{Kappa}   &\textbf{AUC}    & \textbf{F1}   & \textbf{Recall}         \\
\hline
FedAvg                     & 0.7247           & 0.8435 & 0.4552 & 0.4954  \\
FedAvg-$N$                & 0.7227           & 0.8513 & 0.4448 & 0.4944  \\
\textbf{FACL}     & 0.7298           & 0.8451 & 0.4403 & 0.4883  \\
\textbf{FACL-$N$}  & \textbf{0.7342}  & 0.8521 & 0.4376 & 0.4842  \\
\bottomrule
\end{tabular}
\end{table}

The Kappa score exhibited a notable enhancement in the Gleason score, as shown in \textbf{Table \ref{tab_8}}. Across the six centers, the average Kappa score reached 0.6437, whereas the FedAvg achieved a Kappa score of 0.7247. Notably, FedAvg's Kappa outperformed that of the \textit{Centralized} model, underscoring the robustness of models trained through federated learning. 

The comparative results of the FedAvg and FACL models for the test set (Nanchang) are presented in \textbf{Table \ref{tab_9}}. With the integration of an attention-consistent mechanism, the FACL model demonstrated superior performance compared with FedAvg. FACL achieved a Kappa score of 0.7298, surpassing FedAvg's 0.7247, and FACL-$N$ at 0.7342 outperforms FedAvg-$N$ at 0.7227. This emphasizes the effectiveness of the attention-consistent learning mechanism introduced in the FACL model.

\subsection{Interpretability and whole slide attention visualization}

To assess the effectiveness of the proposed model in capturing the morphological features of prostate cancer, a random subset of the WSIs was selected from the test set. Heatmaps, as shown in \textbf{Fig. \ref{Fig5}}, were generated using a patch size of $224 \times 224$ and 90\% overlay. The attention score of each patch was normalized to the range [0, 1], with high scores indicating regions crucial for diagnosis. RGB color maps were used to enhance the clarity, with red representing high attention and blue indicating low attention. These heat maps were overlaid onto the original WSIs.

% =================================================================================
% =========================         V. Conclusion         =========================
% =================================================================================
\section{Conclusion}

We propose a novel FL model, referred to as the FACL, designed for diagnosis and Gleason grading of prostate cancer based on pathological images. We incorporated attention consistency to enhance consistency between the server and client models, resulting in improved robustness and accuracy of the trained models. Additionally, to prevent malicious centers from inferring the data of other centers during the model-training process and to safeguard the privacy of both model builders and data providers, we introduced noise into the model using differential privacy. This approach strikes a balance between privacy protection and performance. Although the FACL has made strides in addressing privacy concerns for large-sized medical data, our work still has some limitations. First, the effectiveness of the model was validated using only pathological prostate images. The performance of the model was further verified by testing WSI datasets containing more diverse cancer types. Second, the number of private medical dataset centers used remained relatively small. Future research aims to extend FL applications to various medical diagnostic tasks in diverse scenarios. Collaboration with medical institutions globally could enhance FL in pathology image tasks and enable intelligent approaches to rare disease diagnosis. Furthermore, acknowledging that our current work was conducted in a simulated environment, we recognize the need to consider communication efficiency in real-world systems. In an FL environment, training data remain distributed among numerous centers, each with potentially unreliable and relatively slow network connections. Overcoming the challenges of privacy, security, and communication efficiency associated with FL is crucial for driving further advancements in the healthcare industry.

\section*{Acknowledgments}
This research is supported in part by Science, Technology and Innovation Commission of Shenzhen Municipality (No. WDZC20200818121348001).

\section*{Ethics approval and consent to participate}
The Research Ethics Committee of The Fourth Hospital of Hebei Medical University, China, approved this study.

\section*{Availability of data and materials}
The private test data during the current study is not publicly available due to restrictions in the ethical permit. PANDA can be accessed at https://www.kaggle.com/c/prostate-cancer-grade-assessment. Diagset can be accessed at https://ai-econsilio.diag.pl/. The computer code of the deep learning model will be available for publishing.

%=================================================================================
%                                     REFERENCES   
%=================================================================================
% \clearpage  % new page
\bibliographystyle{IEEEtran}
\bibliography{Bibliography}

\begin{thebibliography}{10}
\providecommand{\url}[1]{#1}
\csname url@samestyle\endcsname
\providecommand{\newblock}{\relax}
\providecommand{\bibinfo}[2]{#2}
\providecommand{\BIBentrySTDinterwordspacing}{\spaceskip=0pt\relax}
\providecommand{\BIBentryALTinterwordstretchfactor}{4}
\providecommand{\BIBentryALTinterwordspacing}{\spaceskip=\fontdimen2\font plus
\BIBentryALTinterwordstretchfactor\fontdimen3\font minus \fontdimen4\font\relax}
\providecommand{\BIBforeignlanguage}[2]{{%
\expandafter\ifx\csname l@#1\endcsname\relax
\typeout{** WARNING: IEEEtran.bst: No hyphenation pattern has been}%
\typeout{** loaded for the language `#1'. Using the pattern for}%
\typeout{** the default language instead.}%
\else
\language=\csname l@#1\endcsname
\fi
#2}}
\providecommand{\BIBdecl}{\relax}
\BIBdecl

\bibitem{sung2021global}
H.~Sung, J.~Ferlay, R.~L. Siegel, M.~Laversanne, I.~Soerjomataram, A.~Jemal, and F.~Bray, ``Global cancer statistics 2020: Globocan estimates of incidence and mortality worldwide for 36 cancers in 185 countries,'' \emph{CA: a cancer journal for clinicians}, vol.~71, no.~3, pp. 209--249, 2021.

\bibitem{lu2015automated}
C.~Lu and M.~Mandal, ``Automated analysis and diagnosis of skin melanoma on whole slide histopathological images,'' \emph{Pattern Recognition}, vol.~48, no.~8, pp. 2738--2750, 2015.

\bibitem{schirris2022deepsmile}
Y.~Schirris, E.~Gavves, I.~Nederlof, H.~M. Horlings, and J.~Teuwen, ``Deepsmile: Contrastive self-supervised pre-training benefits msi and hrd classification directly from h\&e whole-slide images in colorectal and breast cancer,'' \emph{Medical Image Analysis}, vol.~79, p. 102464, 2022.

\bibitem{chikontwe2022feature}
P.~Chikontwe, S.~J. Nam, H.~Go, M.~Kim, H.~J. Sung, and S.~H. Park, ``Feature re-calibration based multiple instance learning for whole slide image classification,'' in \emph{International Conference on Medical Image Computing and Computer-Assisted Intervention}.\hskip 1em plus 0.5em minus 0.4em\relax Springer, 2022, pp. 420--430.

\bibitem{kartasalo2021artificial}
K.~Kartasalo, W.~Bulten, B.~Delahunt, P.-H.~C. Chen, H.~Pinckaers, H.~Olsson, X.~Ji, N.~Mulliqi, H.~Samaratunga, T.~Tsuzuki \emph{et~al.}, ``Artificial intelligence for diagnosis and gleason grading of prostate cancer in biopsies—current status and next steps,'' \emph{European Urology Focus}, vol.~7, no.~4, pp. 687--691, 2021.

\bibitem{ma2023identification}
X.~Ma, L.~Chen, T.~Chen, K.~Chen, H.~Zhang, K.~Huang, H.~Zheng, H.~Jin, Z.~Cheng, K.~Xiao \emph{et~al.}, ``Identification of a 24-gene panel and a novel marker of podxl2 essential for the pathological diagnosis of early prostate cancer,'' \emph{Computational and Structural Biotechnology Journal}, vol.~21, pp. 5476--5490, 2023.

\bibitem{delahunt2012gleason}
B.~Delahunt, R.~J. Miller, J.~R. Srigley, A.~J. Evans, and H.~Samaratunga, ``Gleason grading: past, present and future,'' \emph{Histopathology}, vol.~60, no.~1, pp. 75--86, 2012.

\bibitem{silva2020going}
J.~Silva-Rodr{\'\i}guez, A.~Colomer, M.~A. Sales, R.~Molina, and V.~Naranjo, ``Going deeper through the gleason scoring scale: An automatic end-to-end system for histology prostate grading and cribriform pattern detection,'' \emph{Computer Methods and Programs in Biomedicine}, vol. 195, p. 105637, 2020.

\bibitem{short2019gleason}
E.~Short, A.~Y. Warren, and M.~Varma, ``Gleason grading of prostate cancer: a pragmatic approach,'' \emph{Diagnostic Histopathology}, vol.~25, no.~10, pp. 371--378, 2019.

\bibitem{GEORGE2022262}
\BIBentryALTinterwordspacing
R.~S. George, A.~Htoo, M.~Cheng, T.~M. Masterson, K.~Huang, N.~Adra, H.~Z. Kaimakliotis, M.~Akgul, and L.~Cheng, ``Artificial intelligence in prostate cancer: Definitions, current research, and future directions,'' \emph{Urologic Oncology: Seminars and Original Investigations}, vol.~40, no.~6, pp. 262--270, 2022. [Online]. Available: \url{https://www.sciencedirect.com/science/article/pii/S1078143922000771}
\BIBentrySTDinterwordspacing

\bibitem{campanella2019clinical}
G.~Campanella, M.~G. Hanna, L.~Geneslaw, A.~Miraflor, V.~Werneck Krauss~Silva, K.~J. Busam, E.~Brogi, V.~E. Reuter, D.~S. Klimstra, and T.~J. Fuchs, ``Clinical-grade computational pathology using weakly supervised deep learning on whole slide images,'' \emph{Nature medicine}, vol.~25, no.~8, pp. 1301--1309, 2019.

\bibitem{wang2022sclwc}
\BIBentryALTinterwordspacing
X.~Wang, J.~Xiang, J.~Zhang, S.~Yang, Z.~Yang, M.-H. Wang, J.~Zhang, Y.~Wei, J.~Huang, and X.~Han, ``{SCL}-{WC}: Cross-slide contrastive learning for weakly-supervised whole-slide image classification,'' in \emph{Advances in Neural Information Processing Systems}, A.~H. Oh, A.~Agarwal, D.~Belgrave, and K.~Cho, Eds., 2022. [Online]. Available: \url{https://openreview.net/forum?id=1fKJLRTUdo}
\BIBentrySTDinterwordspacing

\bibitem{wang2023retccl}
X.~Wang, Y.~Du, S.~Yang, J.~Zhang, M.~Wang, J.~Zhang, W.~Yang, J.~Huang, and X.~Han, ``Retccl: Clustering-guided contrastive learning for whole-slide image retrieval,'' \emph{Medical Image Analysis}, vol.~83, p. 102645, 2023.

\bibitem{xiang2023automatic}
J.~Xiang, X.~Wang, X.~Wang, J.~Zhang, S.~Yang, W.~Yang, X.~Han, and Y.~Liu, ``Automatic diagnosis and grading of prostate cancer with weakly supervised learning on whole slide images,'' \emph{Computers in Biology and Medicine}, vol. 152, p. 106340, 2023.

\bibitem{bulten2022artificial}
W.~Bulten, K.~Kartasalo, P.-H.~C. Chen, P.~Str{\"o}m, H.~Pinckaers, K.~Nagpal, Y.~Cai, D.~F. Steiner, H.~van Boven, R.~Vink \emph{et~al.}, ``Artificial intelligence for diagnosis and gleason grading of prostate cancer: the panda challenge,'' \emph{Nature medicine}, vol.~28, no.~1, pp. 154--163, 2022.

\bibitem{bulten2020automated}
W.~Bulten, H.~Pinckaers, H.~van Boven, R.~Vink, T.~de~Bel, B.~van Ginneken, J.~van~der Laak, C.~Hulsbergen-van~de Kaa, and G.~Litjens, ``Automated deep-learning system for gleason grading of prostate cancer using biopsies: a diagnostic study,'' \emph{The Lancet Oncology}, vol.~21, no.~2, pp. 233--241, 2020.

\bibitem{takahashi2023artificial}
Y.~Takahashi, E.~Dungubat, H.~Kusano, and T.~Fukusato, ``Artificial intelligence and deep learning: New tools for histopathological diagnosis of nonalcoholic fatty liver disease/nonalcoholic steatohepatitis,'' \emph{Computational and Structural Biotechnology Journal}, 2023.

\bibitem{perincheri2021independent}
S.~Perincheri, A.~W. Levi, R.~Celli, P.~Gershkovich, D.~Rimm, J.~S. Morrow, B.~Rothrock, P.~Raciti, D.~Klimstra, and J.~Sinard, ``An independent assessment of an artificial intelligence system for prostate cancer detection shows strong diagnostic accuracy,'' \emph{Modern Pathology}, vol.~34, no.~8, pp. 1588--1595, 2021.

\bibitem{nagpal2020development}
K.~Nagpal, D.~Foote, F.~Tan, Y.~Liu, P.-H.~C. Chen, D.~F. Steiner, N.~Manoj, N.~Olson, J.~L. Smith, A.~Mohtashamian \emph{et~al.}, ``Development and validation of a deep learning algorithm for gleason grading of prostate cancer from biopsy specimens,'' \emph{JAMA oncology}, vol.~6, no.~9, pp. 1372--1380, 2020.

\bibitem{pantanowitz2020artificial}
L.~Pantanowitz, G.~M. Quiroga-Garza, L.~Bien, R.~Heled, D.~Laifenfeld, C.~Linhart, J.~Sandbank, A.~A. Shach, V.~Shalev, M.~Vecsler \emph{et~al.}, ``An artificial intelligence algorithm for prostate cancer diagnosis in whole slide images of core needle biopsies: a blinded clinical validation and deployment study,'' \emph{The Lancet Digital Health}, vol.~2, no.~8, pp. e407--e416, 2020.

\bibitem{mun2021yet}
Y.~Mun, I.~Paik, S.-J. Shin, T.-Y. Kwak, and H.~Chang, ``Yet another automated gleason grading system (yaaggs) by weakly supervised deep learning,'' \emph{npj Digital Medicine}, vol.~4, no.~1, pp. 1--9, 2021.

\bibitem{muti2021development}
H.~S. Muti, L.~R. Heij, G.~Keller, M.~Kohlruss, R.~Langer, B.~Dislich, J.-H. Cheong, Y.-W. Kim, H.~Kim, M.-C. Kook \emph{et~al.}, ``Development and validation of deep learning classifiers to detect epstein-barr virus and microsatellite instability status in gastric cancer: a retrospective multicentre cohort study,'' \emph{The Lancet Digital Health}, vol.~3, no.~10, pp. e654--e664, 2021.

\bibitem{echle2020clinical}
A.~Echle, H.~I. Grabsch, P.~Quirke, P.~A. van~den Brandt, N.~P. West, G.~G. Hutchins, L.~R. Heij, X.~Tan, S.~D. Richman, J.~Krause \emph{et~al.}, ``Clinical-grade detection of microsatellite instability in colorectal tumors by deep learning,'' \emph{Gastroenterology}, vol. 159, no.~4, pp. 1406--1416, 2020.

\bibitem{saldanha2022swarm}
O.~L. Saldanha, P.~Quirke, N.~P. West, J.~A. James, M.~B. Loughrey, H.~I. Grabsch, M.~Salto-Tellez, E.~Alwers, D.~Cifci, N.~Ghaffari~Laleh \emph{et~al.}, ``Swarm learning for decentralized artificial intelligence in cancer histopathology,'' \emph{Nature Medicine}, pp. 1--8, 2022.

\bibitem{pati2022federated}
S.~Pati, U.~Baid, B.~Edwards, M.~Sheller, S.-H. Wang, G.~A. Reina, P.~Foley, A.~Gruzdev, D.~Karkada, C.~Davatzikos \emph{et~al.}, ``Federated learning enables big data for rare cancer boundary detection,'' \emph{Nature communications}, vol.~13, no.~1, pp. 1--17, 2022.

\bibitem{liu2021federated}
Q.~Liu, H.~Yang, Q.~Dou, and P.-A. Heng, ``Federated semi-supervised medical image classification via inter-client relation matching,'' in \emph{International Conference on Medical Image Computing and Computer-Assisted Intervention}.\hskip 1em plus 0.5em minus 0.4em\relax Springer, 2021, pp. 325--335.

\bibitem{li2020federated}
T.~Li, A.~K. Sahu, A.~Talwalkar, and V.~Smith, ``Federated learning: Challenges, methods, and future directions,'' \emph{IEEE Signal Processing Magazine}, vol.~37, no.~3, pp. 50--60, 2020.

\bibitem{xu2021federated}
J.~Xu, B.~S. Glicksberg, C.~Su, P.~Walker, J.~Bian, and F.~Wang, ``Federated learning for healthcare informatics,'' \emph{Journal of Healthcare Informatics Research}, vol.~5, no.~1, pp. 1--19, 2021.

\bibitem{sheller2020federated}
M.~J. Sheller, B.~Edwards, G.~A. Reina, J.~Martin, S.~Pati, A.~Kotrotsou, M.~Milchenko, W.~Xu, D.~Marcus, R.~R. Colen \emph{et~al.}, ``Federated learning in medicine: facilitating multi-institutional collaborations without sharing patient data,'' \emph{Scientific reports}, vol.~10, no.~1, pp. 1--12, 2020.

\bibitem{rieke2020future}
N.~Rieke, J.~Hancox, W.~Li, F.~Milletari, H.~R. Roth, S.~Albarqouni, S.~Bakas, M.~N. Galtier, B.~A. Landman, K.~Maier-Hein \emph{et~al.}, ``The future of digital health with federated learning,'' \emph{NPJ digital medicine}, vol.~3, no.~1, pp. 1--7, 2020.

\bibitem{lu2022federated}
M.~Y. Lu, R.~J. Chen, D.~Kong, J.~Lipkova, R.~Singh, D.~F. Williamson, T.~Y. Chen, and F.~Mahmood, ``Federated learning for computational pathology on gigapixel whole slide images,'' \emph{Medical image analysis}, vol.~76, p. 102298, 2022.

\bibitem{brisimi2018federated}
T.~S. Brisimi, R.~Chen, T.~Mela, A.~Olshevsky, I.~C. Paschalidis, and W.~Shi, ``Federated learning of predictive models from federated electronic health records,'' \emph{International journal of medical informatics}, vol. 112, pp. 59--67, 2018.

\bibitem{boughorbel2019federated}
S.~Boughorbel, F.~Jarray, N.~Venugopal, S.~Moosa, H.~Elhadi, and M.~Makhlouf, ``Federated uncertainty-aware learning for distributed hospital ehr data,'' \emph{arXiv preprint arXiv:1910.12191}, 2019.

\bibitem{huang2019patient}
L.~Huang, A.~L. Shea, H.~Qian, A.~Masurkar, H.~Deng, and D.~Liu, ``Patient clustering improves efficiency of federated machine learning to predict mortality and hospital stay time using distributed electronic medical records,'' \emph{Journal of biomedical informatics}, vol.~99, p. 103291, 2019.

\bibitem{min2019predictive}
X.~Min, B.~Yu, and F.~Wang, ``Predictive modeling of the hospital readmission risk from patients’ claims data using machine learning: a case study on copd,'' \emph{Scientific reports}, vol.~9, no.~1, pp. 1--10, 2019.

\bibitem{duan2020learning}
R.~Duan, M.~R. Boland, Z.~Liu, Y.~Liu, H.~H. Chang, H.~Xu, H.~Chu, C.~H. Schmid, C.~B. Forrest, J.~H. Holmes \emph{et~al.}, ``Learning from electronic health records across multiple sites: A communication-efficient and privacy-preserving distributed algorithm,'' \emph{Journal of the American Medical Informatics Association}, vol.~27, no.~3, pp. 376--385, 2020.

\bibitem{chen2020fedhealth}
Y.~Chen, X.~Qin, J.~Wang, C.~Yu, and W.~Gao, ``Fedhealth: A federated transfer learning framework for wearable healthcare,'' \emph{IEEE Intelligent Systems}, vol.~35, no.~4, pp. 83--93, 2020.

\bibitem{brophy2021estimation}
E.~Brophy, M.~De~Vos, G.~Boylan, and T.~Ward, ``Estimation of continuous blood pressure from ppg via a federated learning approach,'' \emph{Sensors}, vol.~21, no.~18, p. 6311, 2021.

\bibitem{choudhury2019predicting}
O.~Choudhury, Y.~Park, T.~Salonidis, A.~Gkoulalas-Divanis, I.~Sylla \emph{et~al.}, ``Predicting adverse drug reactions on distributed health data using federated learning,'' in \emph{AMIA Annual symposium proceedings}, vol. 2019.\hskip 1em plus 0.5em minus 0.4em\relax American Medical Informatics Association, 2019, p. 313.

\bibitem{chen2021fl}
S.~Chen, D.~Xue, G.~Chuai, Q.~Yang, and Q.~Liu, ``Fl-qsar: a federated learning-based qsar prototype for collaborative drug discovery,'' \emph{Bioinformatics}, vol.~36, no. 22-23, pp. 5492--5498, 2021.

\bibitem{xiong2022facing}
Z.~Xiong, Z.~Cheng, X.~Lin, C.~Xu, X.~Liu, D.~Wang, X.~Luo, Y.~Zhang, H.~Jiang, N.~Qiao \emph{et~al.}, ``Facing small and biased data dilemma in drug discovery with enhanced federated learning approaches,'' \emph{Science China Life Sciences}, vol.~65, no.~3, pp. 529--539, 2022.

\bibitem{yan2021experiments}
B.~Yan, J.~Wang, J.~Cheng, Y.~Zhou, Y.~Zhang, Y.~Yang, L.~Liu, H.~Zhao, C.~Wang, and B.~Liu, ``Experiments of federated learning for covid-19 chest x-ray images,'' in \emph{International Conference on Artificial Intelligence and Security}.\hskip 1em plus 0.5em minus 0.4em\relax Springer, 2021, pp. 41--53.

\bibitem{feki2021federated}
I.~Feki, S.~Ammar, Y.~Kessentini, and K.~Muhammad, ``Federated learning for covid-19 screening from chest x-ray images,'' \emph{Applied Soft Computing}, vol. 106, p. 107330, 2021.

\bibitem{lee2021federated}
H.~Lee, Y.~J. Chai, H.~Joo, K.~Lee, J.~Y. Hwang, S.-M. Kim, K.~Kim, I.-C. Nam, J.~Y. Choi, H.~W. Yu \emph{et~al.}, ``Federated learning for thyroid ultrasound image analysis to protect personal information: Validation study in a real health care environment,'' \emph{JMIR medical informatics}, vol.~9, no.~5, p. e25869, 2021.

\bibitem{dou2021federated}
Q.~Dou, T.~Y. So, M.~Jiang, Q.~Liu, V.~Vardhanabhuti, G.~Kaissis, Z.~Li, W.~Si, H.~H. Lee, K.~Yu \emph{et~al.}, ``Federated deep learning for detecting covid-19 lung abnormalities in ct: a privacy-preserving multinational validation study,'' \emph{NPJ digital medicine}, vol.~4, no.~1, pp. 1--11, 2021.

\bibitem{florescu2022federated}
L.~M. Florescu, C.~T. Streba, M.-S. {\c{S}}erb{\u{a}}nescu, M.~M{\u{a}}muleanu, D.~N. Florescu, R.~V. Teic{\u{a}}, R.~E. Nica, and I.~A. Gheonea, ``Federated learning approach with pre-trained deep learning models for covid-19 detection from unsegmented ct images,'' \emph{Life}, vol.~12, no.~7, p. 958, 2022.

\bibitem{guo2021multi}
P.~Guo, P.~Wang, J.~Zhou, S.~Jiang, and V.~M. Patel, ``Multi-institutional collaborations for improving deep learning-based magnetic resonance image reconstruction using federated learning,'' in \emph{Proceedings of the IEEE/CVF Conference on Computer Vision and Pattern Recognition}, 2021, pp. 2423--2432.

\bibitem{stripelis2021scaling}
D.~Stripelis, J.~L. Ambite, P.~Lam, and P.~Thompson, ``Scaling neuroscience research using federated learning,'' in \emph{2021 IEEE 18th International Symposium on Biomedical Imaging (ISBI)}.\hskip 1em plus 0.5em minus 0.4em\relax IEEE, 2021, pp. 1191--1195.

\bibitem{shiri2021federated}
I.~Shiri, A.~V. Sadr, A.~Sanaat, S.~Ferdowsi, H.~Arabi, and H.~Zaidi, ``Federated learning-based deep learning model for pet attenuation and scatter correction: a multi-center study,'' in \emph{2021 IEEE Nuclear Science Symposium and Medical Imaging Conference (NSS/MIC)}.\hskip 1em plus 0.5em minus 0.4em\relax IEEE, 2021, pp. 1--3.

\bibitem{andreux2020siloed}
M.~Andreux, J.~O.~d. Terrail, C.~Beguier, and E.~W. Tramel, ``Siloed federated learning for multi-centric histopathology datasets,'' in \emph{Domain Adaptation and Representation Transfer, and Distributed and Collaborative Learning}.\hskip 1em plus 0.5em minus 0.4em\relax Springer, 2020, pp. 129--139.

\bibitem{adnan2022federated}
M.~Adnan, S.~Kalra, J.~C. Cresswell, G.~W. Taylor, and H.~R. Tizhoosh, ``Federated learning and differential privacy for medical image analysis,'' \emph{Scientific reports}, vol.~12, no.~1, pp. 1--10, 2022.

\bibitem{wang2023generalizable}
X.~Wang, J.~Zhang, S.~Yang, J.~Xiang, F.~Luo, M.~Wang, J.~Zhang, W.~Yang, J.~Huang, and X.~Han, ``A generalizable and robust deep learning algorithm for mitosis detection in multicenter breast histopathological images,'' \emph{Medical Image Analysis}, vol.~84, p. 102703, 2023.

\bibitem{baid2022federated}
U.~Baid, S.~Pati, T.~M. Kurc, R.~Gupta, E.~Bremer, S.~Abousamra, S.~P. Thakur, J.~H. Saltz, and S.~Bakas, ``Federated learning for the classification of tumor infiltrating lymphocytes,'' \emph{arXiv preprint arXiv:2203.16622}, 2022.

\bibitem{ogier2023federated}
J.~Ogier~du Terrail, A.~Leopold, C.~Joly, C.~B{\'e}guier, M.~Andreux, C.~Maussion, B.~Schmauch, E.~W. Tramel, E.~Bendjebbar, M.~Zaslavskiy \emph{et~al.}, ``Federated learning for predicting histological response to neoadjuvant chemotherapy in triple-negative breast cancer,'' \emph{Nature Medicine}, pp. 1--12, 2023.

\bibitem{kairouz2021advances}
P.~Kairouz, H.~B. McMahan, B.~Avent, A.~Bellet, M.~Bennis, A.~N. Bhagoji, K.~Bonawitz, Z.~Charles, G.~Cormode, R.~Cummings \emph{et~al.}, ``Advances and open problems in federated learning,'' \emph{Foundations and Trends{\textregistered} in Machine Learning}, vol.~14, no. 1--2, pp. 1--210, 2021.

\bibitem{de2022mitigating}
A.~B. de~Luca, G.~Zhang, X.~Chen, and Y.~Yu, ``Mitigating data heterogeneity in federated learning with data augmentation,'' \emph{arXiv preprint arXiv:2206.09979}, 2022.

\bibitem{ahn2022communication}
S.~Ahn, S.~Kim, Y.~Kwon, J.~Park, J.~Youn, and S.~Cho, ``Communication-efficient diffusion strategy for performance improvement of federated learning with non-iid data,'' \emph{arXiv preprint arXiv:2207.07493}, 2022.

\bibitem{nguyen2022federated}
H.~Nguyen, L.~Phan, H.~Warrier, and Y.~Gupta, ``Federated learning for non-iid data via client variance reduction and adaptive server update,'' \emph{arXiv preprint arXiv:2207.08391}, 2022.

\bibitem{mcmahan2017communication}
B.~McMahan, E.~Moore, D.~Ramage, S.~Hampson, and B.~A. y~Arcas, ``Communication-efficient learning of deep networks from decentralized data,'' in \emph{Artificial intelligence and statistics}.\hskip 1em plus 0.5em minus 0.4em\relax PMLR, 2017, pp. 1273--1282.

\bibitem{li2021fedrs}
X.-C. Li and D.-C. Zhan, ``Fedrs: Federated learning with restricted softmax for label distribution non-iid data,'' in \emph{Proceedings of the 27th ACM SIGKDD Conference on Knowledge Discovery \& Data Mining}, 2021, pp. 995--1005.

\bibitem{gao2022feddc}
L.~Gao, H.~Fu, L.~Li, Y.~Chen, M.~Xu, and C.-Z. Xu, ``Feddc: Federated learning with non-iid data via local drift decoupling and correction,'' in \emph{Proceedings of the IEEE/CVF Conference on Computer Vision and Pattern Recognition}, 2022, pp. 10\,112--10\,121.

\bibitem{hanzely2021personalized}
F.~Hanzely, B.~Zhao, and M.~Kolar, ``Personalized federated learning: A unified framework and universal optimization techniques,'' \emph{arXiv preprint arXiv:2102.09743}, 2021.

\bibitem{wu2022adaptive}
X.~Wu, Y.~Zhang, M.~Shi, P.~Li, R.~Li, and N.~N. Xiong, ``An adaptive federated learning scheme with differential privacy preserving,'' \emph{Future Generation Computer Systems}, vol. 127, pp. 362--372, 2022.

\bibitem{wang2022safeguarding}
C.~Wang, X.~Wu, G.~Liu, T.~Deng, K.~Peng, and S.~Wan, ``Safeguarding cross-silo federated learning with local differential privacy,'' \emph{Digital Communications and Networks}, vol.~8, no.~4, pp. 446--454, 2022.

\bibitem{dwork2006our}
C.~Dwork, K.~Kenthapadi, F.~McSherry, I.~Mironov, and M.~Naor, ``Our data, ourselves: Privacy via distributed noise generation,'' in \emph{Annual international conference on the theory and applications of cryptographic techniques}.\hskip 1em plus 0.5em minus 0.4em\relax Springer, 2006, pp. 486--503.

\bibitem{zhao2022cork}
J.~Zhao, H.~Zhu, F.~Wang, R.~Lu, H.~Li, J.~Tu, and J.~Shen, ``Cork: A privacy-preserving and lossless federated learning scheme for deep neural network,'' \emph{Information Sciences}, vol. 603, pp. 190--209, 2022.

\bibitem{li2021model}
Q.~Li, B.~He, and D.~Song, ``Model-contrastive federated learning,'' in \emph{Proceedings of the IEEE/CVF Conference on Computer Vision and Pattern Recognition}, 2021, pp. 10\,713--10\,722.

\bibitem{litian2020federated}
T.~Li, A.~K. Sahu, M.~Zaheer, M.~Sanjabi, A.~Talwalkar, and V.~Smith, ``Federated optimization in heterogeneous networks,'' \emph{Proceedings of Machine Learning and Systems}, vol.~2, pp. 429--450, 2020.

\bibitem{malaviya2023fedfame}
S.~Malaviya, M.~Shukla, P.~Korat, and S.~Lodha, ``Fedfame: A data augmentation free framework based on model contrastive learning for federated semi-supervised learning,'' in \emph{Proceedings of the 38th ACM/SIGAPP Symposium on Applied Computing}, 2023, pp. 1114--1121.

\bibitem{hosseini2023proportionally}
S.~M. Hosseini, M.~Sikaroudi, M.~Babaie, and H.~Tizhoosh, ``Proportionally fair hospital collaborations in federated learning of histopathology images,'' \emph{IEEE Transactions on Medical Imaging}, 2023.

\bibitem{hosseini2022cluster}
S.~M. Hosseini, M.~Sikaroudi, M.~Babaei, and H.~R. Tizhoosh, ``Cluster based secure multi-party computation in federated learning for histopathology images,'' in \emph{International Workshop on Distributed, Collaborative, and Federated Learning, Workshop on Affordable Healthcare and AI for Resource Diverse Global Health}.\hskip 1em plus 0.5em minus 0.4em\relax Springer, 2022, pp. 110--118.

\bibitem{wang2022transformer}
X.~Wang, S.~Yang, J.~Zhang, M.~Wang, J.~Zhang, W.~Yang, J.~Huang, and X.~Han, ``Transformer-based unsupervised contrastive learning for histopathological image classification,'' \emph{Medical Image Analysis}, vol.~81, p. 102559, 2022.

\bibitem{ilse2018attention}
M.~Ilse, J.~Tomczak, and M.~Welling, ``Attention-based deep multiple instance learning,'' in \emph{International conference on machine learning}.\hskip 1em plus 0.5em minus 0.4em\relax PMLR, 2018, pp. 2127--2136.

\bibitem{wang2023sac}
X.~Wang, D.~Cai, S.~Yang, Y.~Cui, J.~Zhu, K.~Wang, and J.~Zhao, ``Sac-net: Enhancing spatiotemporal aggregation in cervical histological image classification via label-efficient weakly supervised learning,'' \emph{IEEE Transactions on Circuits and Systems for Video Technology}, 2023.

\bibitem{4310076}
N.~Otsu, ``A threshold selection method from gray-level histograms,'' \emph{IEEE Transactions on Systems, Man, and Cybernetics}, vol.~9, no.~1, pp. 62--66, 1979.

\bibitem{joyce2011kullback}
J.~M. Joyce, ``Kullback-leibler divergence,'' in \emph{International encyclopedia of statistical science}.\hskip 1em plus 0.5em minus 0.4em\relax Springer, 2011, pp. 720--722.

\bibitem{ziller2021medical}
A.~Ziller, D.~Usynin, R.~Braren, M.~Makowski, D.~Rueckert, and G.~Kaissis, ``Medical imaging deep learning with differential privacy,'' \emph{Scientific Reports}, vol.~11, no.~1, p. 13524, 2021.

\bibitem{yang2019federated}
Q.~Yang, Y.~Liu, T.~Chen, and Y.~Tong, ``Federated machine learning: Concept and applications,'' \emph{ACM Transactions on Intelligent Systems and Technology (TIST)}, vol.~10, no.~2, pp. 1--19, 2019.

\end{thebibliography}
\biboptions{sort&compress}
%=================================================================================
%                                    Appendix     
%=================================================================================
\clearpage % new page
% \onecolumn
\section*{Appendix}
\begin{table}[h]
		\setlength{\abovecaptionskip}{0.02cm}   
		\setlength{\belowcaptionskip}{0cm}
\setcounter{table}{0}
\renewcommand{\thetable}{S\arabic{table}} 
\caption{\label{tab_s1} The diagnostic model's performance on DiagSet-A was reported as the five-fold mean}
\centering
{
\begin{tabular}{llllll} 
\toprule
$\alpha$ &  \makecell[l]{\textbf{Training} \\ \textbf{settings}} &\textbf{AUC}  & \textbf{F1}  & \textbf{ACC}   & \textbf{Recall}   \\ \hline
                    & Hebei-1     & 0.9664 & 0.8888 & 0.8899 & 0.8281  \\
/                   & Hebei-2     & 0.9539 & 0.8742 & 0.8524 & 0.9647  \\
                    & Nanchang     & 0.9544 & 0.7483 & 0.6440 & 0.9955  \\ 
\hline
\multirow{6}*{0.05} & DiagSet-B-1   & 0.9579 & 0.8134 & 0.8313 & 0.6916  \\
~                   & DiagSet-B-2     & 0.9628 & 0.8888 & 0.8899 & 0.8281  \\
~                   & PANDA-1       & 0.9474 & 0.8741 & 0.8711 & 0.8414  \\
~                   & PANDA-2     & 0.9596 & 0.6964 & 0.5386 & 0.9955  \\
~                   & FedAvg         & 0.9635 & 0.9079 & 0.8969 & 0.9559  \\
~                   & Centralized             & 0.9670 & 0.9082 & 0.9063 & 0.8722  \\ 
\hline 
\multirow{6}*{0.1}  & DiagSet-B-1 & 0.9605 & 0.8940 & 0.8922 & 0.8546  \\
~                   & DiagSet-B-2       & 0.9637 & 0.8737 & 0.8782 & 0.7929  \\
~                   & PANDA-1          & 0.9498 & 0.9126 & 0.9063 & 0.9207  \\
~                   & PANDA-2         & 0.9572 & 0.8712 & 0.8477 & 0.9691  \\
~                   & FedAvg            & 0.9648 & 0.9204 & 0.9133 & 0.9427  \\
~                   & Centralized   & 0.9631   & 0.8619    & 0.8665  & 0.7841                 \\ 
\hline
\multirow{6}*{0.3}  & DiagSet-B-1     & 0.9663     & 0.9062    & 0.9016     & 0.8942    \\
~                   & DiagSet-B-2       & 0.9684     & 0.8803    & 0.8829     & 0.8105    \\
~                   & PANDA-1         & 0.9541     & 0.8457    & 0.8103     & 0.9779    \\
~                   & PANDA-2        & 0.9614     & 0.8912    & 0.8782     & 0.9383    \\
~                   & FedAvg           & 0.9651     & 0.8972    & 0.8852     & 0.9427   \\
~                   & Centralized      & 0.9712     & 0.9162    & 0.9110     & 0.9162  \\ 
\hline 
\multirow{6}*{0.5} &  DiagSet-B-1     & 0.9655   & 0.9027    & 0.9016   & 0.8590         \\
~                  &  DiagSet-B-2     & 0.9686   & 0.9045    & 0.9016   & 0.8766         \\
~                  &  PANDA-1         & 0.9488   & 0.7174    & 0.5831   & 0.9955         \\
~                  &  PANDA-2         & 0.9617   & 0.9193    & 0.9133   & 0.9295         \\
~                  &  FedAvg           & 0.9647   & 0.8957    & 0.8805   & 0.9647         \\
~                  &  Centralized      & 0.9697   & 0.9115    & 0.9086   & 0.8854         \\
\bottomrule
\end{tabular}
}
\end{table}
\begin{table}[h]
		\setlength{\abovecaptionskip}{-8.575cm}  
		\setlength{\belowcaptionskip}{0cm}
\setcounter{table}{1}
\renewcommand{\thetable}{S\arabic{table}} 
\caption{\label{tab_s2} The diagnostic model's performance on QHD was reported as the five-fold mean}
\centering
{
\begin{tabular}{llllll} 
\toprule
$\alpha$ &  \makecell[l]{\textbf{Training} \\ \textbf{settings}}  &\textbf{AUC}   & \textbf{F1}  & \textbf{ACC}   & \textbf{Recall}    \\
\hline
&  Hebei-1      & 0.9636       & 0.9204      & 0.9150           & 0.8683        \\
/ &  Hebei-2      & 0.9805       & 0.8914      & 0.8653           & 0.9769        \\
&  Nanchang      & 0.9731       & 0.8201      & 0.7568           & 0.9792        \\ 
\hline
\multirow{6}*{0.05} &  DiagSet-B-1 & 0.9727 & 0.8274  & 0.8326 & 0.7090  \\
~ &  DiagSet-B-2     & 0.9734       & 0.8769       & 0.8745        & 0.7898   \\
~ &  PANDA-1        & 0.9769       & 0.8259       & 0.8313       & 0.7066  \\
~ &  PANDA-2       & 0.9838       & 0.7234       & 0.5673       & 1.0000     \\
~ & FedAvg                  &  0.9795              & 0.9439       & 0.9346           & 0.9722         \\
~ & Centralized          &  0.9788            & 0.9467   & 0.9385    & 0.9653       \\ 
\hline  
\multirow{6}*{0.1} &  DiagSet-B-1     & 0.9718     & 0.9268      & 0.9215      & 0.8775      \\
~ &  DiagSet-B-2       & 0.9736     & 0.8866      & 0.8836      & 0.8036       \\
~ &  PANDA-1          & 0.9766     & 0.9407      & 0.9333      & 0.9353      \\
~ &  PANDA-2         & 0.9778     & 0.7487      & 0.6209      & 0.9976       \\
~ & FedAvg            & 0.9775     & 0.9426      & 0.9333      & 0.9676                \\
~ & Centralized          & 0.9731            & 0.9477      & 0.9398    & 0.9630       \\ 
\hline  
\multirow{6}*{0.3}&  DiagSet-B-1  & 0.9761  & 0.9247  & 0.9202 & 0.8661  \\
~ &  DiagSet-B-2       & 0.9773     & 0.8953       & 0.8915      & 0.8198     \\
~ &  PANDA-1          & 0.9727     & 0.8782       & 0.8471      & 0.9745     \\
~ &  PANDA-2        & 0.9785     & 0.9442       & 0.9359      & 0.9584     \\
~ & FedAvg            & 0.9788     & 0.9523       & 0.9451      & 0.9699                 \\
~ & Centralized        & 0.9802       & 0.9501       & 0.9424      & 0.9676               \\ 
\hline
\multirow{6}*{0.5}&  DiagSet-B-1 & 0.9761  & 0.9129  & 0.9084  & 0.8475  \\
~ &  DiagSet-B-2      & 0.9765     & 0.9323      & 0.9267    & 0.8914            \\
~ &  PANDA-1   & 0.9722        & 0.7386      & 0.6013    & 0.9953  \\
~ &  PANDA-2      & 0.9776        & 0.9512      & 0.9450    & 0.9468               \\
~ &  FedAvg       &   0.9765        & 0.9522      & 0.9451    & 0.9676             \\
~ & Centralized     & 0.9766     & 0.9532    & 0.9464   & 0.9653                     \\
\bottomrule
\end{tabular}
}
\end{table}
\end{document}